\documentclass{article}
\pdfoutput=1

     \usepackage[preprint]{neurips_2020}

\usepackage[utf8]{inputenc} 
\usepackage[T1]{fontenc}    
\usepackage{hyperref}       
\usepackage{url}            
\usepackage{booktabs}       
\usepackage{amsfonts}       
\usepackage{nicefrac}       
\usepackage{microtype}      
\usepackage{bm}
\usepackage{amsmath}
\usepackage{amsthm}
\usepackage{graphicx}
\usepackage{subcaption}
\usepackage{verbatim}
\usepackage{wrapfig}
\usepackage{cleveref}
\usepackage{multirow}
\usepackage{mathtools}

\usepackage[table]{xcolor}
\usepackage{collcell}
\usepackage{hhline}
\usepackage{pgf}

\newcolumntype{E}{>{\collectcell\ColCell}c<{\endcollectcell}}

\usepackage[textsize=footnotesize,color=green!40]{todonotes}
\usepackage{footnote}
\makesavenoteenv{tabular}
\usepackage{tabularx}

\makesavenoteenv{table}

\newcommand{\norm}[1]{\left\lVert#1\right\rVert}

\usepackage{amsmath,amssymb,commath,bm}
\usepackage[linesnumbered,ruled,vlined]{algorithm2e}

\let\oldnl\nl
\newcommand{\nonl}{\renewcommand{\nl}{\let\nl\oldnl}}

\DeclareMathOperator*{\minimize}{minimize}

\theoremstyle{definition}
\newtheorem{definition}{Definition}[section]

\usepackage{amsmath,amsfonts,bm}

\def\eqref#1{equation~\ref{#1}}

\def\1{\bm{1}}

\DeclareMathAlphabet{\mathsfit}{\encodingdefault}{\sfdefault}{m}{sl}
\SetMathAlphabet{\mathsfit}{bold}{\encodingdefault}{\sfdefault}{bx}{n}

\DeclarePairedDelimiter\parentheses{\lparen}{\rparen}
\newcommand{\br}[1]{\parentheses*{#1}}

\DeclarePairedDelimiter\squareparentheses{\lbrack}{\rbrack}
\newcommand{\sqbr}[1]{\squareparentheses*{#1}}

\DeclareMathOperator*{\Exp}{E}
\newcommand{\ExP}[2]{\Exp_{#2}\sqbr{#1}}

\newcommand{\namewal}{ReSWAT }

\title{Towards transformation-resilient provenance detection of digital media}

\author{
	Jamie Hayes
\quad\quad\quad
	Krishnamurthy (Dj) Dvijotham
\quad\quad\quad
	Yutian Chen
\\ \\
\textbf{Sander Dieleman} 
\quad\quad\quad
\textbf{Pushmeet Kohli}
\quad\quad\quad
\textbf{Norman Casagrande}
	\\ \\
	DeepMind
	\\
	\texttt{jamhay@google.com}
}

\begin{document}

\maketitle

\begin{abstract}
Advancements in deep generative models have made it possible to synthesize images, videos and audio signals that are difficult to distinguish from natural signals, creating opportunities for potential abuse of these capabilities. This motivates the problem of tracking the provenance of signals, i.e., being able to determine the original source of a signal. Watermarking the signal at the time of signal creation is a potential solution, but current techniques are brittle and watermark detection mechanisms can easily be bypassed by applying post-processing transformations (cropping images, shifting pitch in the audio etc.). In this paper, we introduce {\bf \namewal (Resilient Signal Watermarking via Adversarial Training)}, a framework for learning transformation-resilient watermark detectors that are able to detect a watermark even after a signal has been through several post-processing transformations. Our detection method can be applied to  domains with continuous data representations such as images, videos or sound signals. Experiments on watermarking image and audio signals show that our method can reliably detect the provenance of a signal, even if it has been through several post-processing transformations, and improve upon related work in this setting. Furthermore, we show that for specific kinds of transformations (perturbations bounded in the $\ell_2$ norm), we can even get formal guarantees on the ability of our model to detect the watermark.
We provide qualitative examples of watermarked image and audio samples in \url{https://drive.google.com/open?id=1-yZ0WIGNu2Iez7UpXBjtjVgZu3jJjFga}.
\end{abstract}

\section{Introduction}

Generative models have contributed to impressive advancements in content generation
and representation learning in both digital image and audio domains \citep{brock2018large, kalchbrenner2018efficient, mehri2016samplernn, zhu2017unpaired, prenger2019waveglow, donahue2019large, oord2016wavenet, goodfellow2014generative, kingma2013auto}. However, as generative models learn to better
match a target distribution, the distinction between natural signals and synthetic signals
generated by a model has blurred, leading to a raft of concerns over the potential misuse of these models \citep{chesney2018deep}. For example, synthetic videos that are indistinguishable from natural videos, sometimes referred to as \emph{deep fakes}, have the potential to cause widespread distrust in traditional media. 

While the problem of detecting synthetic signals is interesting in its own right, it is challenging to do in a manner that is independent of the model used to generate the signal. Instead, we consider the problem of provenance detection via watermarking, a technique that involves injection of a carefully chosen (but imperceptible) perturbation (watermark) into the signal at the time of creation. The presence of the perturbation in the signal can be later used to detect the provenance (or ultimate source) of this signal. 
While the simplicity of the technique makes watermarking an appealing technique for detecting synthetic (as well as naturally generated) signals, it is susceptible to adversarial actors that can  systematically try to remove the watermark by transforming the signal (so as to obfuscate the provenance). In the case of images, this may take the form of cropping, addition of Gaussian noise or blurring, rotating images etc. Many watermarking schemes break down under these types of transformations.

In this paper, we propose a novel  \emph{transformation-resilient watermarking} scheme -- Resilient Signal Watermarking via Adversarial Training ( \namewal) that is able to detect the presence of a watermark even after the signal has been through systematic  attempts to remove the watermark. We use ideas from the literature on adversarial training \citep{madry2017towards} to learn to synthesize watermarks (encoding the provenance of a signal) that can be detected even after the watermarked signal has been through post-processing transformations. Our detection method can operate in any domain with continuous data representations. Experimental results demonstrate that our method is effective and substantially improves over existing methods - it can reliably detect the provenance of a synthetic signal even if the signal has been deformed or manipulated. 

\namewal is designed to protect against the abuse of digital media released by an individual or organization. For example, consider an organization that owns a corpus of image/audio/video samples released to the public. These samples could be distorted and/or used to create deep fakes by a malicious party. If the organization has watermarked these samples with our scheme, since the watermark is designed to be robust to post-processing transformations, the watermark will persist through transformations introduced by the malicious party. The organization can then verify if the deep fake was originally created using samples that they own.

\subsection{Contributions}
\begin{itemize}
    \item[1] We formulate the transformation resilient watermarking problem and show how it can be reduced to an empirical risk minimization problem with a minimax loss function.
    \item[2] We develop a transformation resilient watermarking scheme, named \namewal (Resilient Signal Watermarking via Adversarial Training) and show that the learning procedure is closely related to adversarial training techniques \citep{madry2017towards, athalye2017synthesizing}. 
    \item[3] Empirically, we show across a set of image and audio datasets that our scheme can produce imperceptible watermarks that can be detected even after the signal has been through a series of adversarial transformations that preserve fidelity to the original signal. We do this using both standard metrics for signal watermarking and via human evaluation on audio signals.
    Furthermore, we give formal guarantees of watermark detection if an adversarial
    transformation has a bounded $\ell_2$ norm.
\end{itemize}

\subsection{Related work}\label{sec:related_work}

The problem of detecting synthetic signals has recently received a significant amount of attention. \cite{marra2018gans} investigate how a generative 
adversarial network (GAN) \citep{goodfellow2014generative} leaves an identifiable fingerprint in the image it generates, while \cite{nataraj2019detecting} detect if an image was generated by a GAN by training a detection model on co-occurence matrices of the RGB channels of images. Similarly, \cite{yu2018attributing} also train a model that learns to detect if an image was generated by a GAN, and attribute a generated image to its source model. These methods are specifically designed to operate in the image domain and it is unknown how resilient they are to transformations that an attacker could apply.

Digital watermarking consists of a two stage process: An embedding stage, where the original signal is combined with a hidden message (also referred to as the \emph{watermark}), producing a watermarked version of the signal. Secondly, a \emph{decoding} stage, where the hidden message is retrieved from the watermarked signal. Watermarking schemes are typically evaluated on fidelity of the watermarked signal to the original signal, the resilience of the watermark to various post-processing transformations and the false positive rates (i.e. whether the decoder detects watermarks in non-watermarked signals). Robustness of watermarking techniques is measured against a number of practical attacks that aim to remove or degrade the watermark signal. For example, adding Gaussian noise, image cropping and image compression represent watermark degrading, watermark removal and watermark geometric (re-positioning) attacks, respectively. 

\subsection{Comparison to related work}\label{sec:comparison_related_work}

In this paper, we are primarily interested in zero-bit watermarking, where one is simply interested in detecting the presence of a watermark (rather than decoding a hidden message from the watermark). The state-of-the-art zero-bit watermarking scheme in the image domain is referred to as \emph{Broken Arrows} (BA) \citep{furon2008broken} and won the ``Break Our Watermarking System'' (BOWS) competition \citep{bennour2007watermarking}. BA has a provable minimum false positive rate under Gaussian noise, however the scheme does not provide robustness
against geometric attacks (like rotations and cropping). To the best of our knowledge, no publicly available zero-bit watermarking scheme has been published that claims superiority over BA, and it is still used as a baseline in contemporary related work (c.f. section 4.1 in~\cite{quiring2018forgotten}). We compare \namewal with BA in \cref{ssec:image_eval2} and \cref{sec:imagenet_full_acc_complete_psnr}. 

Recent work has used deep learning to develop \emph{multi-bit} watermarking schemes \citep{mun2017robust, ahmadi2018redmark, zhu2018hidden, zhang2019robust, wen2019romark, luo2020distortion}; we discuss and compare relevant approaches in \cref{sec:comparison_multi_bit_related_work}.

\section{Formulation of the robust watermarking problem}
\label{sec:problem_defintion_threat_model}

We study signals that live in a space $\mathcal{X}$ and are generated by a probabilistic source $P_s$ - our framework applies regardless of whether $P_s$ is a natural source (images of natural scenes) or an artificial source (samples from a VAE or a GAN). We are interested in developing a scheme to watermark signals  generated by this source, with the requirement that: 

\begin{itemize}
    \item[1] The detector should detect a watermark in any watermarked signal from source $P_s$ and not detect a watermark in any other signal.
    \item[2] Even if post-processing transformations (for example, compression/cropping/frequency shift etc) are applied to the signal, the detector detects the watermark, or the absence thereof. 
\end{itemize}

Formally, we define the robust watermarking problem as follows: 
\begin{definition}[Transformation resilient watermarking scheme]
Consider a watermarking scheme $(W, D)$ where $W: \mathcal{X} \mapsto \mathcal{X}$ is a watermarking routine and $D: \mathcal{X} \mapsto \{0, 1\}$ is a watermark detector. Let $\mathcal{T}$ be a space of transformations with $T: \mathcal{X} \mapsto \mathcal{X}$ for each $T\in\mathcal{T}$. We consider the watermarking scheme resilient with respect to a set of transformations $\mathcal{T}$ for a source $P_s$ if
$ D\br{T\br{W\br{s}}}=1, D\br{T\br{s}} =0, \forall T \in \mathcal{T}$
  with high probability for $s \sim P_s$. 
\end{definition}

This formulation suggests an empirical risk minimization approach for training $D$. Given a fixed $W$, we can train $D$ to minimize the empirical risk

\vspace{-0.5cm}

\begin{align}
\ExP{\max_{T \in \mathcal{T}} \ell\br{D\br{T\br{W\br{s}}}, 1} + \ell\br{D\br{T\br{s}}, 0}}{s \sim P_s}
\end{align}

where $\ell$ is a loss function measuring the discrepancy between the prediction of $D$ and the desired label ($0$ for non-watermarked signals and $1$ for watermarked signals).

The inner maximization transformations resembles the objective used in adversarial training \citep{madry2017towards}, motivating our approach to learning a transformation resilient watermarking. We develop this idea in the following section.
 
\section{\namewal: Resilient watermarking via adversarial training}
\label{sec:watermarking_scheme}

We parameterize the detector $D$ as a neural network $f_{\theta}$, parameterized by $\theta$. If we fix the detector, the watermark embedding process $W$ should embed a watermark that is ``strongly detected'' by the detector, i.e., the watermark embedding mechanism pushes the watermark deep inside the decision boundary of $D$. However, we also want the watermark to be imperceptible so we require that the watermark does not significantly change the original signal according to some distance measure. While it is challenging to define the space of imperceptible distortions of the input, a convenient proxy is to limit the change in terms of the $\ell_\infty$ norm between the watermarked and the original signal. Thus, we construct the watermark, $\delta$, by solving the following optimization problem:

\vspace{-0.5cm}

\begin{align}
W(s) = \min_{\norm{\delta}_\infty\leq \epsilon} \ell(f_\theta(s+\delta), 1)
\end{align}

\vspace{-0.3cm}

This optimization problem can be solved efficiently using a projected gradient descent (PGD) method, and is mathematically similar to computing an adversarial example for a neural network (however, in this case, the adversary is ``friendly'' and actually pushes the example to minimize the loss of the detector). The parameter $\epsilon$ controls the perceptibility of the watermark. If $\epsilon$ is too large, the watermark will be clearly perceptible while if it is too small, the watermark may be easily washed out by post-processing steps. Thus, the right choice of $\epsilon$ achieves the optimal trade-off between perceptibility and transformation resilient detection. Note that we use the $\ell_{\infty}$ norm of the perturbation instead of another norm since
this distributes the watermark over the entire signal, and so will be resilient to geometric attacks (that black out patches of the input, or obfuscate inputs).

With this watermark embedding method, we can now train the detector to perform transformation resilient watermark detection as follows:

\vspace{-0.5cm}

\begin{align}
   \minimize_{\theta, \norm{\delta}_{\infty}\leq \epsilon} \ExP{\max_{T \in \mathcal{T}} \ell\br{f_\theta\br{T\br{s + \delta}}, 1} + \ell\br{f_\theta\br{T\br{s}}, 0}}{s \sim P_s} \label{eq2}
\end{align}

\vspace{-0.3cm}

This is similar to the expectation over transformation attack \citep{athalye2017synthesizing}, that aims
to construct adversarial examples that persist through a set of transformations. During training, we only 
sample from differentiable transformations, this is so we do not have to approximate the gradient, 
which would be prohibitively slow to training. 

In practice, at each iteration in training we approximate finding
$\max_{T\in\mathcal{T}}$ with $\max_{\{T_1, T_2, ..., T_n\}}$, where $T_i\in\mathcal{T}, i\in\{1,...,n\}$. Theoretically,
if we sample more transformations we achieve a better approximation for the true gradient over the full transformation distribution, however, empirically,
we found it too expensive to trade more parallel transformations for longer training time. We explore how resilience to transformations is affected
by the number of sampled transformations at each step in training in \cref{sec:trans_sample_exp}. 

\subsection{Optimizing for non-transferable watermarks}

Traditional watermarking schemes are based on a secret key that ensures that only parties with access to the key can create watermarked content; an attacker that can watermark content without access to the key represents an integrity attack on the scheme. We refer to this kind of attack as a \emph{specificity} attack that attempts to introduce false positives. In our case, the classifier  detecting/constructing the watermark can be thought of as an approximation of a secret key. 

We will now propose a method to construct watermark perturbations for a given watermark classifier, $f_{\theta}$, which do not transfer to other watermark classifiers trained on the same data.

Recall that given a signal $s$, to create a watermark perturbation we use PGD which takes steps in the direction of
\begin{equation}
    -\nabla_s \max_{T \in \mathcal{T}} \Big[ \ell \big(f_{\theta}(T(s)), 1\big)\Big] \label{eq_standard}
\end{equation}
To encourage that the generated perturbation doesn't transfer to other watermark classifiers trained on the same data, we generate the watermark that is robust against an ensemble of models. Let $\{ f_{\theta}^i: \mathcal{X}\rightarrow \{0,1\}, i\in\{1,...,n\} \}$ be an ensemble of watermark classifiers trained on the same problem as $f_\theta$. To limit the transferability of watermarks on $f_\theta$, we create watermark perturbations by stepping in the direction of

\begin{equation}
    -\bigg( \nabla_s \max_{T\in \mathcal{T}} \Big[ \ell \big(f_\theta(T(s)), 1\big)\Big]  + \nabla_s \max_{i, T\in \mathcal{T}} \Big[ \ell \big(f_{\theta}^i(T(s)), 0\big)\Big] \bigg) \label{eq4}
\end{equation}

Taking steps in the direction of this loss will increase the probability that
watermark perturbations are only classified as watermarked by $f_\theta$ and do not transfer
to other models. 
In experiments, we take $\ell(\cdot, \cdot)$ to be binary cross-entropy loss.

\section{Evaluation}
\label{sec:eval}

Here, we give experimental results that our scheme is robust to a variety of attacks. We evaluate the scheme on image data in \cref{ssec:image_eval2}, and evaluate on audio data in \cref{ssec:audio_eval_appendix}. 

\subsection{Evaluating performance on image data}
\label{ssec:image_eval2}

\begin{table}[]
\centering
\caption{Transformations used to evaluate the robustness of \namewal in the image space.}
\label{tab:image_transformations}
\resizebox{0.9\columnwidth}{!}{%
\begin{tabular}{cc}
\toprule
         Transform & Transform description \\
\cmidrule{1-2}
Gaussian noise  & Add Gaussian noise with standard deviation $\sigma$.                                                    \\
\addlinespace[0.15cm]
  
  Rotation &      Rotate an image by a random angle uniformly sampled from range [$-r, r$].                      \\
  
\addlinespace[0.15cm]
  \multirow{2}{*}{Cropping} & Randomly crop an image's height by $c_h$ pixels and width by $c_w$ pixels, \\ &
   and project back to the original image size. \\
   
  \addlinespace[0.15cm]
  
  Horizontal flip & With probability $\nicefrac{1}{2}$, flip the image on the horizontal axis. \\
  
  \addlinespace[0.15cm]
  \multirow{2}{*}{Brightness} & Randomly adjust the brightness of an image by a factor, $b$,
  which ranges  \\ & from 0 (no increase in brightness) to 1 (maximum whitening).\\

  \bottomrule
\end{tabular}%
}
\end{table}

During training we sample from a set of transformations, to which the scheme
should be robust. In this evaluation we use transformations
detailed in \cref{tab:image_transformations}.
We evaluate watermark detection robustness in the image space on both Cifar10 \citep{cifar10} and ImageNet \citep{imagenet_cvpr09}. We defer a full evaluation of Cifar10 to \cref{sec:eval_on_cifar10}.  

\begin{figure*}[t!]
    \centering
    \begin{subfigure}[t]{0.31\textwidth}
        \centering
        \includegraphics[width=1.0\textwidth]{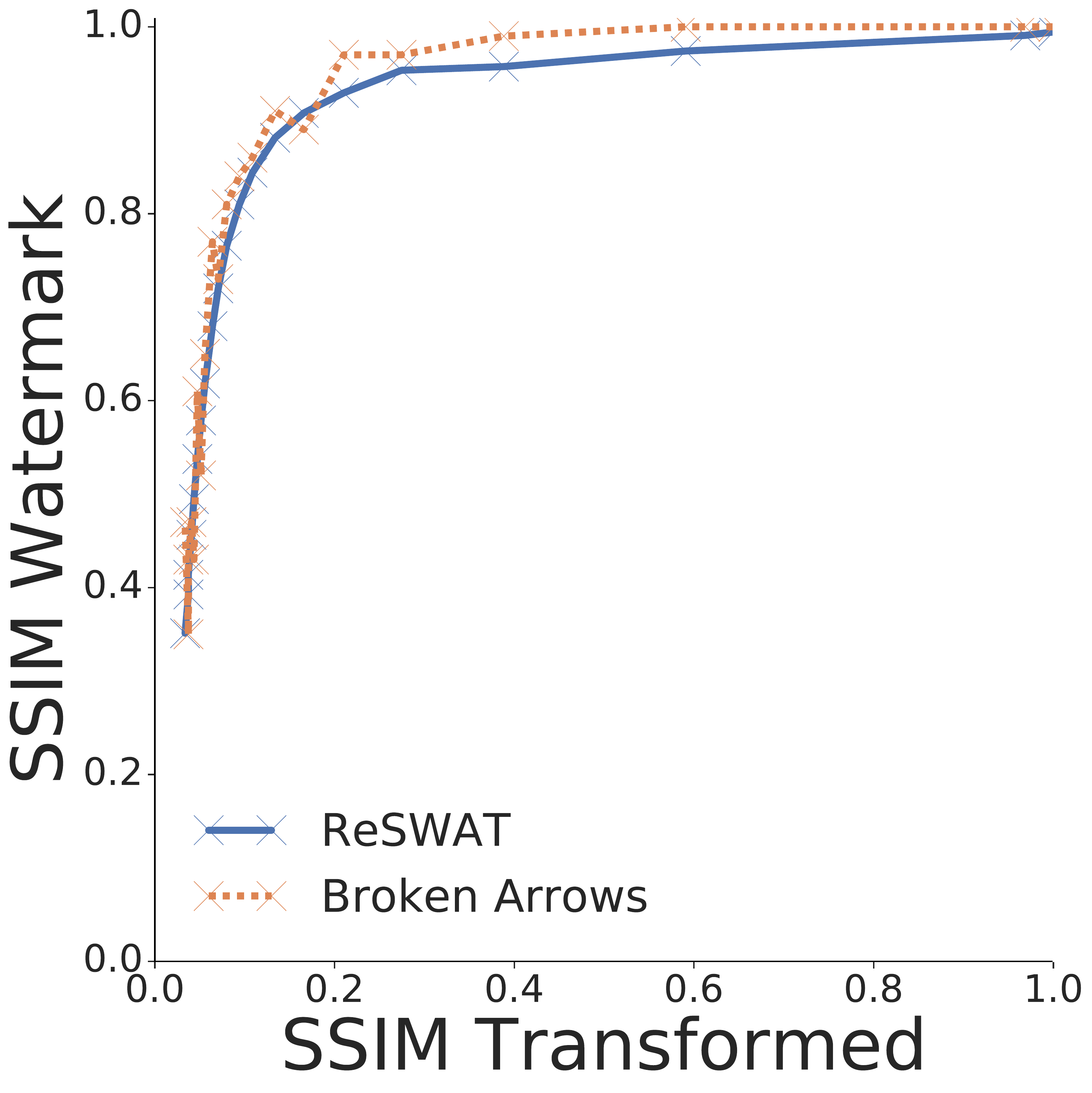}
        \label{fig:imagenet_full_acc_ssim_gaussian}
        \caption{Gaussian noise.}
    \end{subfigure}
    \begin{subfigure}[t]{0.31\textwidth}
        \centering
        \includegraphics[width=1.0\textwidth]{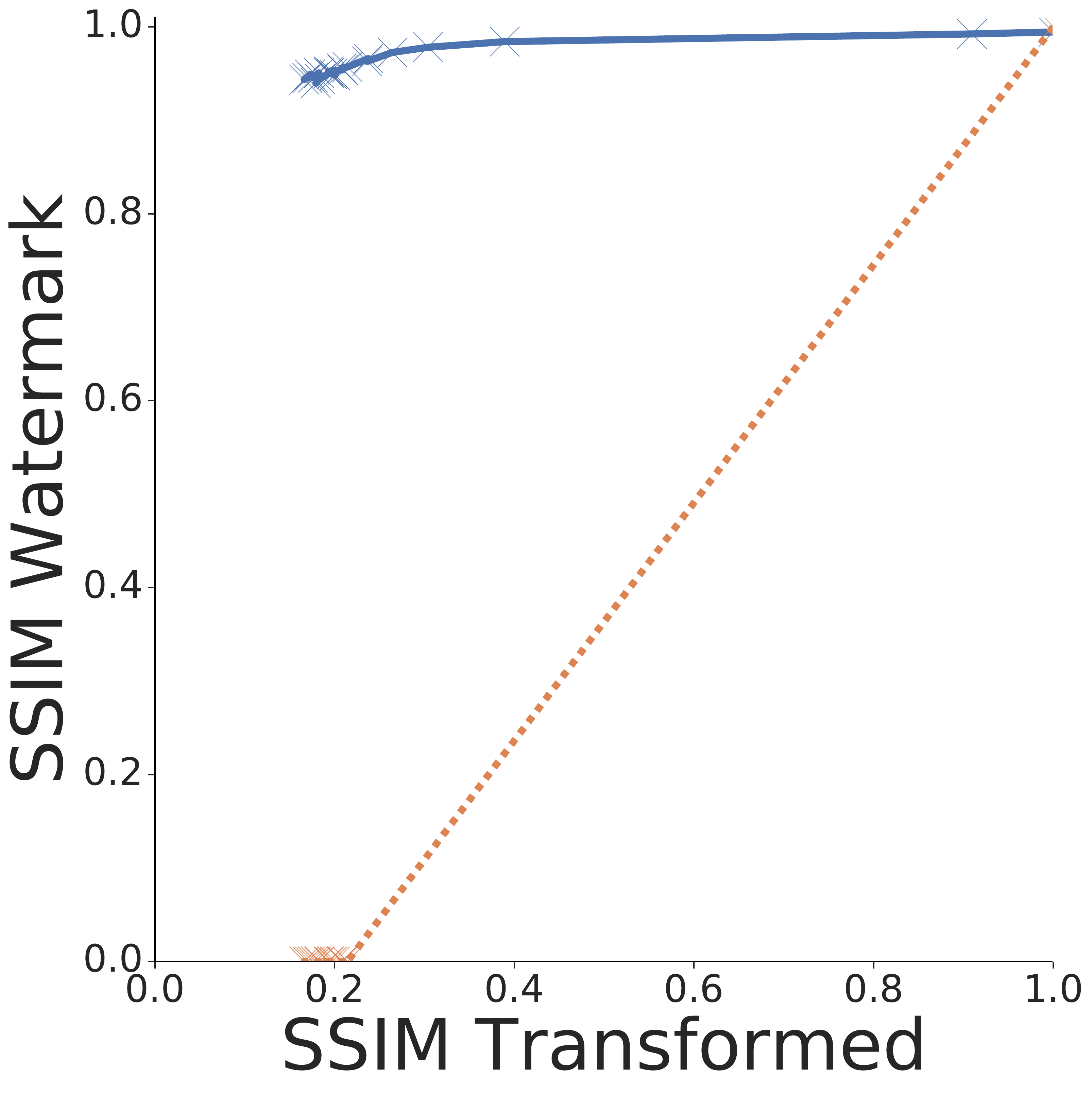}
        \label{fig:imagenet_full_acc_ssim_rotation}
        \caption{Rotation.}
    \end{subfigure}
    \begin{subfigure}[t]{0.31\textwidth}
        \centering
        \includegraphics[width=1.0\textwidth]{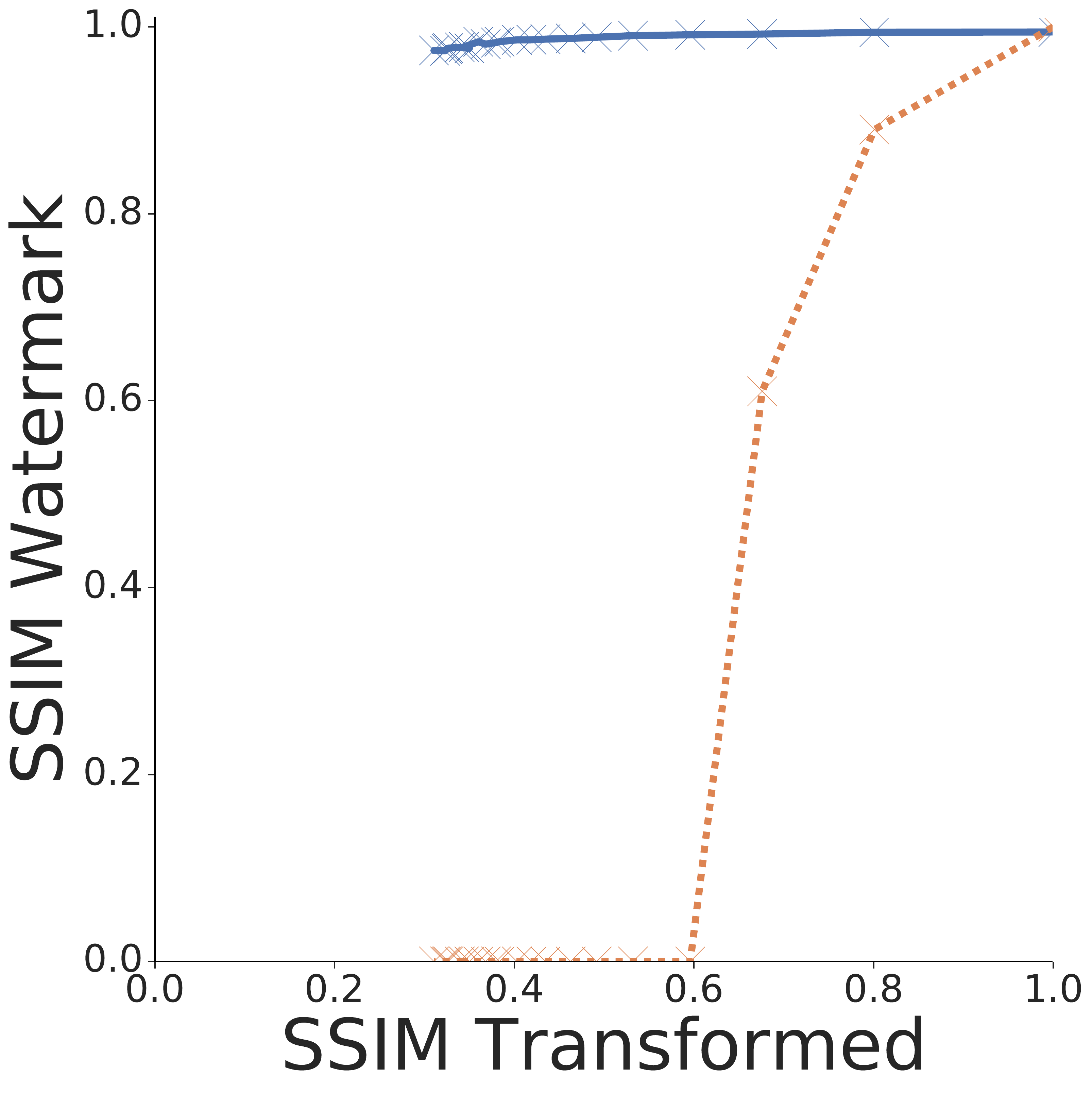}
        \label{fig:imagenet_full_acc_ssim_cropping}
        \caption{Crop.}
    \end{subfigure}
    \caption{Distortion needed for perfect provenance detection (by $f_{\text{comp}}^{\epsilon_{10}}$) against different signal quality degradation levels suffered by the attacker. We measure the amount of distortion
    the watermarking scheme must introduce in order to guarantee perfect detection accuracy
    under a transformation, and compare with the Broken Arrows zero-bit watermarking scheme.}
   \label{fig:imagenet_full_acc_complete_ssim}
\end{figure*}

\noindent \textbf{ImageNet.} We train a standard ResNet152 classifier \citep{he2016deep}, modified for binary prediction, for the watermark detector. We replace batch normalization with instance
normalization \citep{ulyanov2016instance}, as 
we noticed that batch statistics were heavily skewed by the transformations applied during training. We train with a mini-batch size of 32 for 60,000 steps, with an initial learning rate of 0.1 and decaying this by a factor of 10 every 20,000 steps. 

All images are normalized into the range [0,1], and a batch of watermarked images are constructed with five PGD steps at each iteration. With respect to the set of transformations, during training, we set $\sigma$ to 0.25,
$r$ to $\nicefrac{\pi}{2}$, $c_h$ and $c_w$ to 10, and $b$ to 0.1. We trained four classifiers under these 
hyperparameters, two where at each step we randomly sample a single transformation, and  watermark with $\epsilon=\nicefrac{5}{255}$ and $\epsilon=\nicefrac{10}{255}$, referred to as 
$f^{\epsilon_5}$ and $f^{\epsilon_{10}}$, respectively, and two where we apply a composition of all transformations, and watermark with $\epsilon=\nicefrac{5}{255}$ and $\epsilon=\nicefrac{10}{255}$, 
referred to as
$f_{\text{comp}}^{\epsilon_5}$ and $f_{\text{comp}}^{\epsilon_{10}}$, respectively. 

All watermark detection classifiers achieved $100\%$ test set accuracy, where the test set consists of 10,000 non-watermarked and 10,000 watermarked
ImageNet test set images. A detailed comparison of the differences between these models is given in \cref{sec:eval_on_cifar10}, however, we found that models trained under a composition of 
transformations are more robust to all attacks and so remaining experiments will use only
$f_{\text{comp}}^{\epsilon_5}$ and $f_{\text{comp}}^{\epsilon_{10}}$. For evaluation,  
we measure the success of an attack with respect 
to the distortion introduced by the attack measured by  
structural similarity (SSIM) score \citep{wang2004image}.

We now describe experiments measuring the quality of our watermarking scheme under various transformations seeking to induce a misclassification in the watermark detector. We study both attacks that seek to remove a watermark from given watermarked signals (thus inducing a false negative for the detector) and attacks that seek to make the detector detect a watermark in a non-watermarked image (thus inducing a false positive).

\paragraph{Signal transformation attacks (false negatives).}

We investigate the trade-off in distortion of the watermarked
image when we require perfect detection under various transformations. Given a watermarked signal $s$, we sample from a space of transformations to create $n$ copies of this input under, referred to as $s^i$, $i\in\{1,..., n\}$. We increase the watermark $\epsilon$ value until the accuracy of the detection model on $s^i$, $i\in\{1,..., n\}$ is >99\%. We
measure the perceptibility of the watermark and the perceptibility of the transformation
used in the attack, both in terms of SSIM, and compare with Broken Arrows (BA), where for each transformation we set the number of random samples, $n$, to 1 million. 

\Cref{fig:imagenet_full_acc_complete_ssim} shows that for various distributions of transformations
the amount of distortion introduced by the watermarking scheme is dominated by
the amount of distortion introduced by the transformation. For Gaussian noise, the necessary distortion introduced by the watermark (for perfect detection under a transformation) using \namewal is approximately
equivalent to BA and substantially smaller than the distortion introduced by the 
transformation. While for cropping and rotation attacks, our watermarks incur negligible levels of
distortion while BA fails to watermark content without incurring large distortions. We show qualitative examples
in \cref{fig:imagenet_full_acc_noise_examples} of \cref{sec:eval_on_cifar10} for a Gaussian noise attack, and analogous plots of \cref{fig:imagenet_full_acc_complete_ssim} using Peak-Signal-to-Noise ratio as the distortion metric in \cref{sec:imagenet_full_acc_complete_psnr}.

\paragraph{Specificity attacks (false positives).}

\begin{figure*}[t!]
\captionsetup{width=0.45\textwidth}
\centering
\begin{minipage}[t]{.49\linewidth}
  \centering
  \includegraphics[width=\linewidth]{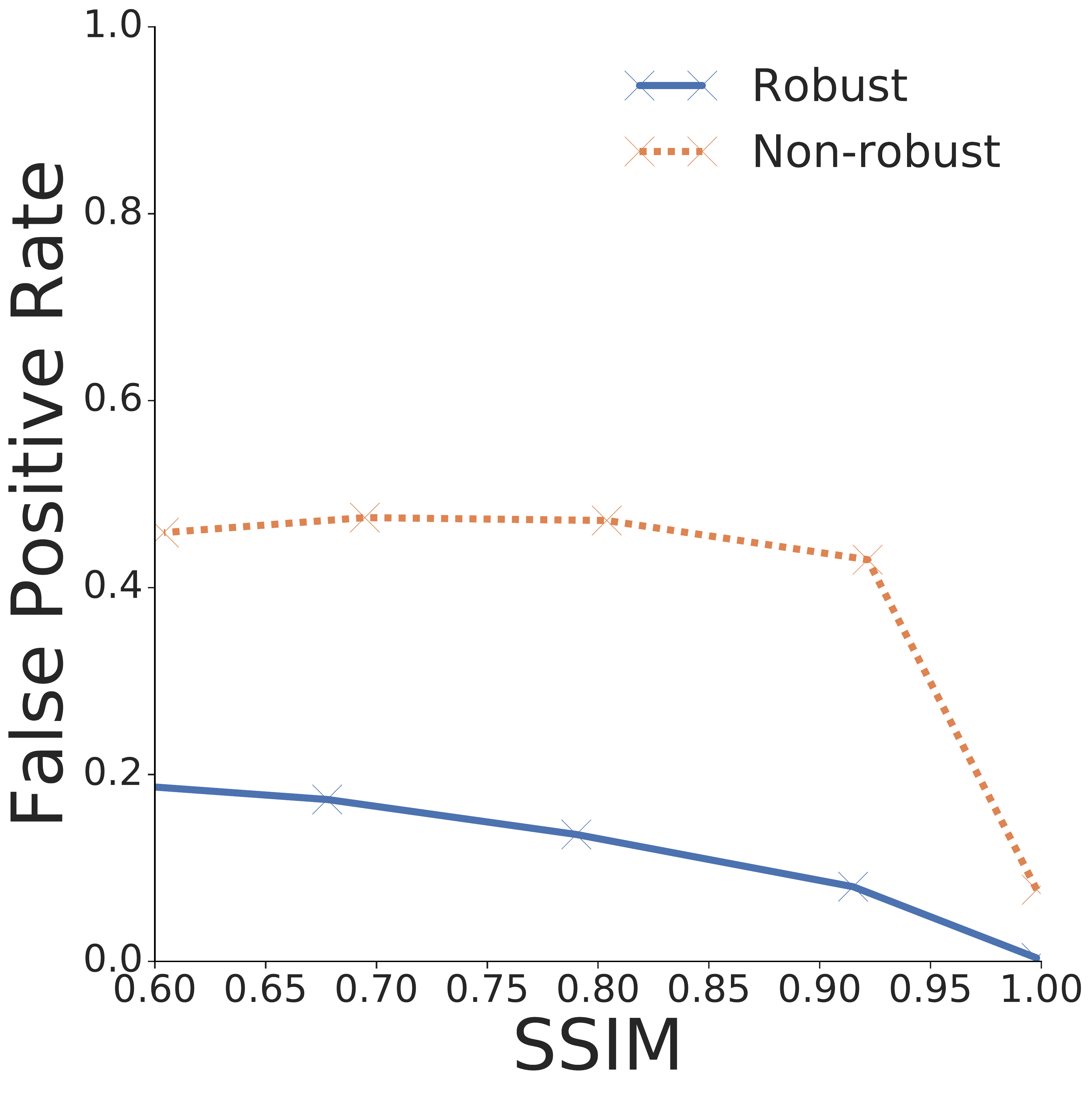}
  \caption{Average false positive rate of $f_{\text{comp}}^{\epsilon_{5}}$ (non-robust) and $\hat{f}_{\text{comp}}^{\epsilon_{5}}$ (robust) on watermark's constructed by other watermark classifiers. Given as a function of the average watermark image SSIM.}
  \label{fig:robust_transfer_imagenet}
\end{minipage}%
\begin{minipage}[t]{0.49\linewidth}
  \centering
  \includegraphics[width=\linewidth]{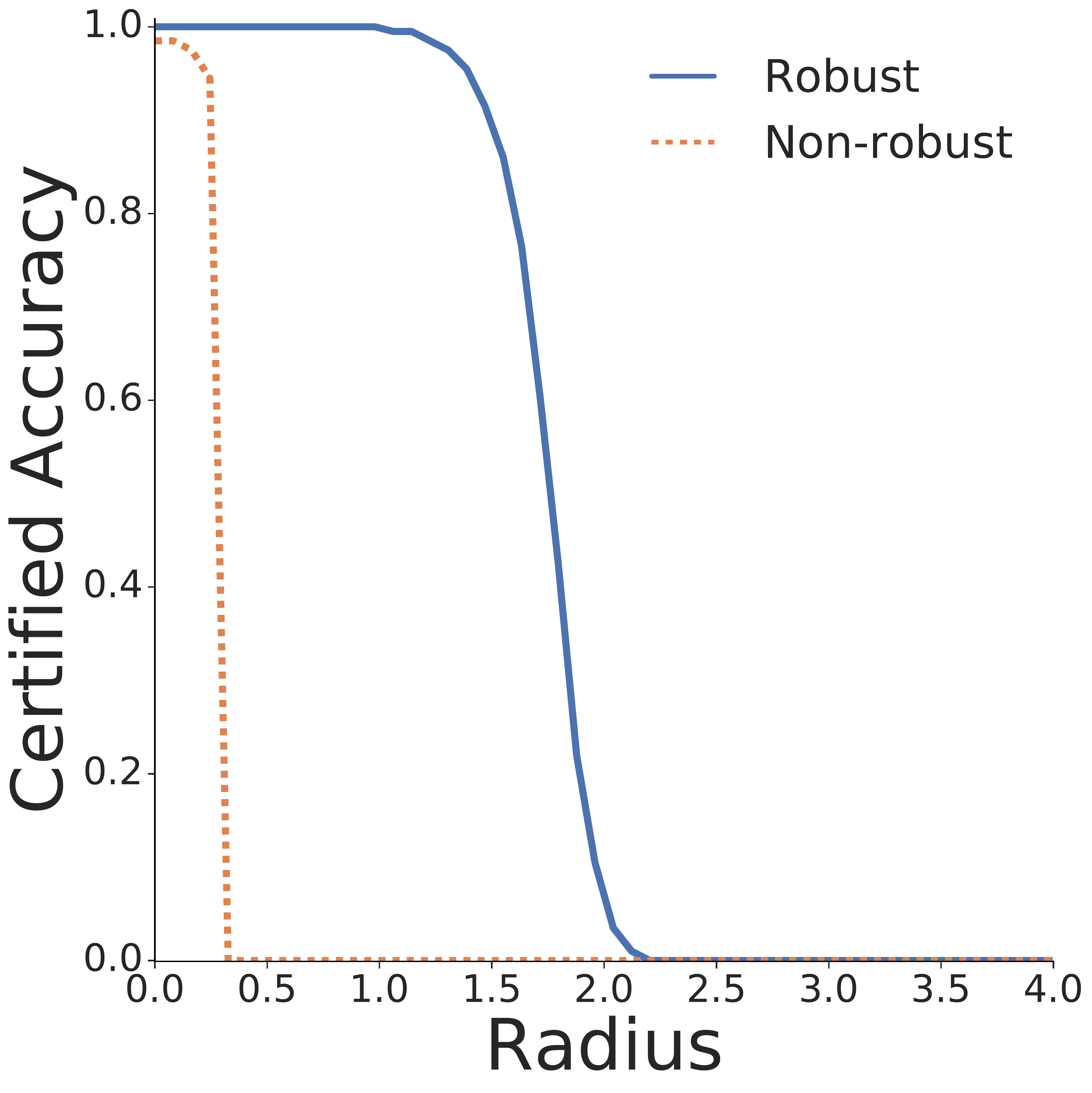}
        \caption{Certified accuracy of 200 watermarked ImageNet examples. 
        Plotted as a function of the $\ell_2$ radius of robustness. We compare with robustness guarantees against a non-robust watermark detection model, that has not been trained on a set of transformations.}
   \label{fig:randomized_smoothing}
\end{minipage}
\end{figure*}

Using the ImageNet dataset, we trained a model with the 
same hyperparameters as $f_{\text{comp}}^{\epsilon_{5}}$,
however the model was optimized by constructing watermarks using \cref{eq4} instead of \cref{eq_standard}, using five other pre-trained watermark classifiers~\footnote{The five other pre-trained watermark classifiers are five versions of $f_{\text{comp}}^{\epsilon_{5}}$ trained with different random initializations.}; we  denote this model by  $\hat{f}_{\text{{comp}}}^{\epsilon_{5}}$. Given 20 pre-trained watermark classifiers only differing from 
$f_{\text{{comp}}}^{\epsilon_{5}}$ by random initialization of weights, we compare how well 1,000 watermark's constructed using each of these models transfers to $f_{\text{{comp}}}^{\epsilon_{5}}$ and
$\hat{f}_{\text{{comp}}}^{\epsilon_{5}}$, representing an attack that attempts to introduce false positives. 
\Cref{fig:robust_transfer_imagenet} shows the difference in average false positive rate between $f_{\text{{comp}}}^{\epsilon_{5}}$ and
$\hat{f}_{\text{{comp}}}^{\epsilon_{5}}$.  
At a watermark SSIM score of 0.60, 
$\hat{f}_{\text{{comp}}}^{\epsilon_{5}}$ has a false positive rate of below 20\% while the false positive rate on these inputs is nearly 50\% for $f_{\text{{comp}}}^{\epsilon_{5}}$. Clearly, optimizing \cref{eq4} improves resilience to specificity attacks. 

\paragraph{Certified robustness.}

The previous sets of experiments present results on the robustness of the watermarking scheme against ``best effort'' attacks, i.e, we attempt to compute attacks that break the watermarking system and declare success if we fail to do so. However, it is possible that our attack algorithm failed to find the worst case transformation that would break the watermark detection. Thus, it is desirable to have a stronger guarantee against all attacks within a certain transformation space.

While provable guarantees are hard to obtain in general, we can leverage recent work on randomized smoothing techniques \citep{cohen2019certified} that are able to obtain certified guarantees against transformations constrained in the $\ell_2$ norm, i.e., the adversary can transform the signal by any amount within a distance $\epsilon$ in the $\ell_2$ norm. On 200 examples from the ImageNet test set that are watermarked with $\epsilon=\nicefrac{20}{255}$, we estimate a lower bound, $\underline{p}$, 
on the probability of the most-likely
class under Gaussian noise parameterized by $\mathcal{N}(0, \sigma^2I)$, using 10,000 random samples. Given $\underline{p}$,
the detector is robust to adversarial perturbations, $\gamma$, 
if $\norm{\gamma}_{2}<\sigma\Phi^{-1}(\underline{p})$ (as proven in \citet{cohen2019certified}), up to a confidence
level of $\alpha$ which we set to 0.99. \Cref{fig:randomized_smoothing} shows
the certified accuracy of these examples as a function of the certified radius in the $\ell_2$-norm. For $\epsilon=\nicefrac{20}{255}$ nearly all 200 watermarked images are robust
to any perturbation
with an $\ell_2$-norm smaller than 1.5. We also include results on a non-robust watermark detection model - a model trained without sampling from a distribution of transformations. This model has a comparatively small certified region of robustness, implying that training on a distribution of transformations does improve robustness.

\begin{table}[t]
\captionsetup{width=1\columnwidth}
\caption{Transformations used to evaluate the robustness of \namewal in the audio space. Transformation values specify the amount of distortion that can be added to an audio sample while maintaining perfect detection accuracy under a trained \namewal model. All audio samples are normalized into [$-\nicefrac{1}{2}$, $\nicefrac{1}{2}$] range.}
\label{tab:tts_transforms}
\centering
\resizebox{\columnwidth}{!}{%
\begin{tabular}{lccccc}
\toprule

& \multicolumn{5}{c}{\makebox[0pt]{Transform description}}                \\
\cmidrule{2-6}
Gaussian noise & \multicolumn{5}{c}{\makebox[0pt]{Add Gaussian noise with standard deviation $\sigma$.}} \\
\addlinespace[0.15cm]
Uniform noise & \multicolumn{5}{c}{\makebox[0pt]{Add uniform noise sampled from a range [$-\alpha,\alpha$].}} \\
\addlinespace[0.15cm]
    
    \multirow{2}{*}{Pitch shift} & \multicolumn{5}{c}{Pitch-shifts audio by a randomly sampled}\\ 
    & \multicolumn{5}{c}{scale factor in range $[-\beta, \beta]$ (Implementation adapted from \cite{Ellis02}).} \\
\addlinespace[0.15cm]
    
Silence & \multicolumn{5}{c}{\makebox[0pt]{Randomly silence a contiguous fraction, $\gamma$, of the audio.}} \\
\addlinespace[0.15cm]
\multirow{2}{*}{Roll} & \multicolumn{5}{c}{Rolls values in audio by a fraction of audio length}\\
& \multicolumn{5}{c}{uniformly sampled at random from range [$-\delta, \delta$].} \\

\cmidrule{1-6}

Input                                       & \multicolumn{5}{c}{\makebox[0pt]{Transform}}                \\
\cmidrule{2-6}
& Gaussian noise ($\sigma=$) & Uniform noise ($\alpha=$) & Pitch shift ($\beta=$) & Silence ($\gamma=$) & Roll ($\delta=$) \\
\cmidrule{2-6}
Watermarked ($\epsilon = 4\times10^{-4}$)   & 0.01              & 0.0056            & 0.16  & 0.78    & 1.0 \\
Watermarked ($\epsilon = 4.8\times10^{-3}$) & 0.013             & 0.0278            & 0.5   & 0.78    & 1.0 \\

\bottomrule
\end{tabular}%
}
\end{table}

\paragraph{Out-of-distribution inputs.}

The previous section focused on attacks on signals drawn from the same distribution as the model was trained on. A natural question to ask is, would the system fail when signals are drawn from a different distribution? To evaluate this, we verified
that, given a detector trained on Cifar10 images,
all SVHN test images are correctly classified to
the non-watermarked class. Secondly, we watermarked 500 images generated by BigGAN \citep{brock2018large}
using a detector trained on ImageNet samples ($f_{\text{{comp}}}^{\epsilon_{10}}$). We then measured resilience to transformations
as detailed in \emph{signal transformation attack (false negatives)} experiments -- we attacked both non-watermarked and watermarked images with Gaussian noise, rotation and cropping such that the average SSIM of these images is 0.6. In comparison, the average SSIM of watermarked images was 0.82. The false positive rate (attacking non-watermarked images) was 1.2\% and the false negative rate (attacking watermarked images) was 0\%. BigGAN watermarked images exhibited comparative levels of resilience to transformations
despite the model being trained on ImageNet samples.
Accompanying experiments on out-of-distribution \emph{transformations} an attacker may apply are given in \cref{sec:ood_transformations}.

\subsection{Evaluation on text-to-speech dataset}
\label{ssec:audio_eval_appendix}

To evaluate \namewal on audio data, we train a watermark detection model using a proprietary dataset composed of hours of high quality speech data, where each audio sample
is a short speech
utterance lasting between 1 and 10 seconds. We use 
a DeepSpeaker architecture \citep{li2017deep}, modified for binary prediction.
The pre-processing stage takes as input, a variable length waveform, normalizes
values between $-\nicefrac{1}{2}$ and $\nicefrac{1}{2}$, 
and outputs the fixed length mel-spectrogram which is then used as input to the model. 
We use a differentiable approximation of the mel-spectrogram so the watermarks can be embedded on the raw waveforms.
We train the detector model for 100,000 steps, with an exponentially decaying learning rate 
initialized at 0.001 and decayed by a factor of 0.9 every 1000 steps. The watermarking
$\epsilon$ value is initialized at 0.04 and decreased by 0.00001 whenever the detection rate is 100\% based on a moving average of the previous 100 steps. We use five PGD steps at each iteration to create watermarked inputs. During training we sample from a set of transformations, described in \cref{tab:tts_transforms}, to which the scheme should be robust. 
At test time we achieve perfect detection rate on 200 watermarked audio samples using
$\epsilon=4\times 10^{-4}$. Similar to the methodology described in \cref{ssec:image_eval2},
we measure the robustness of watermarked audio samples with respect to the 
amount of distortion introduced by transformations, under the requirement that
detection accuracy is >99\%. For a given transformation and input,
we randomly sample the input 100 times under the transformation, this creates a test set of 20,000 data points on which we measure
the accuracy of the detector. \Cref{tab:tts_transforms} shows the results of attacking watermarked
inputs for different $\epsilon$ values using various transformations. Watermarks at $\epsilon = 4\times10^{-4}$ are almost imperceptible, while a faint 
background noise can be heard for $\epsilon = 4.8\times10^{-3}$. The detection model is robust to
a large number of transformations, for example, 78\% of audio can be obscured without decreasing detection accuracy.

\begin{wrapfigure}{r}{0.4\textwidth}
\centering
  \includegraphics[width=\linewidth]{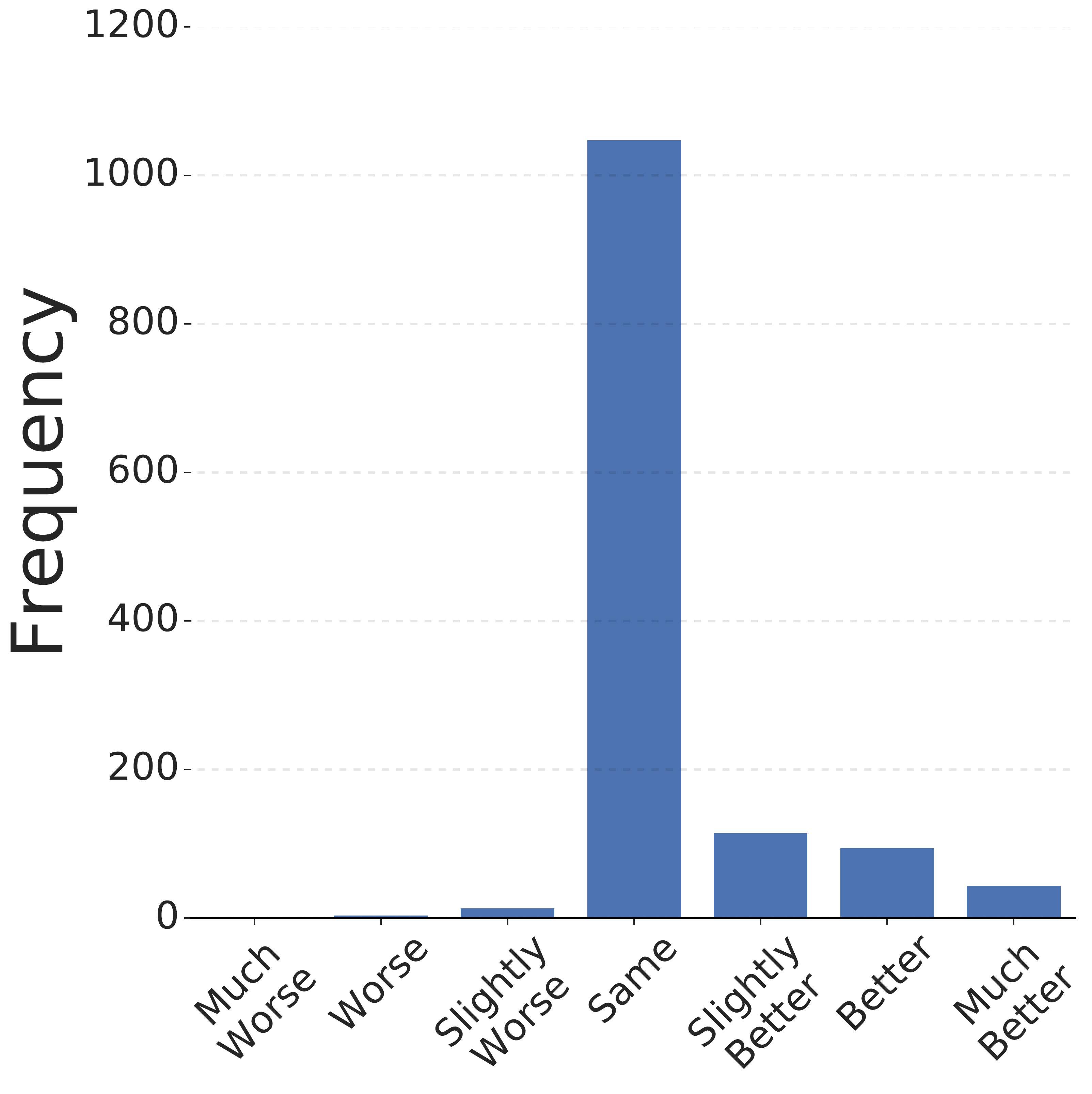}
        \caption{A/B experiment. 660 non-watermarked and watermarked audio samples presented to 1320 people in total. Participants were played the original sample and watermarked sample and asked to rate how different the original audio was from the watermarked version.}
\label{fig:ab_tts}
\end{wrapfigure}

\noindent \textbf{Human evaluation.} Here, we conduct human 
evaluation of watermark perceptibility using both mean opinion score (MOS) and A/B tests. 
There were 2000 unique participants for the MOS test and over 600 for the A/B experiments. 
MOS \citep{streijl2016mean} is a commonly used measure for audio quality; it is expressed as a single rational number, typically in the range 1–5, where 1 is lowest perceived quality, and 5 is the highest perceived quality. 
The MOS is calculated as the arithmetic mean over single ratings performed by human subjects. Due to the large number of audio samples used in the study, we use a single rating per audio sample. Our dataset consisted of 2000 watermarked and non-watermarked audio samples, that were all correctly classified by the watermark detector model under transformations listed in \cref{tab:tts_transforms}. 
The average rating of non-watermarked content was 4.595$\pm$0.576, and the average rating of watermarked content (at $\epsilon = 4\times10^{-4}$)  was 4.530$\pm$0.530. Clearly, human participants did not perceive the watermarked audio as significantly worse or degraded.
For A/B tests we took a subset of 660 watermarked and non-watermarked audio samples. We played both the non-watermarked and watermarked audio sample to participants and asked if the watermarked audio sample was worse or better than the original. Results are shown in 
\cref{fig:ab_tts}; over 80\% of human participants rated the quality of watermarked audio samples as the same as non-watermarked audio samples.

\section{Discussion \& limitations}
\label{sec:limitations}

Our work adheres to common assumptions in zero-bit watermarking research. 
We do not consider an attacker that has access to labeled content containing both non-watermarked and watermarked versions. This is a relatively common assumption in zero-bit watermarking, as if an attacker has a procedure to identify if a piece of content was or was not watermarked they have practically already broken the scheme.
Furthermore, we assume an attacker cannot access arbitrary amounts of watermarked content. In potential practical applications an organization can limit an attacker's exposure to watermarked content by prohibiting any single party from making a large number of queries to access content. Our experiments in \cref{ssec:image_eval2} on false positive attacks are designed to show that an attacker that does not have this information, but has other information about the watermarking procedure, cannot break the scheme. The attacker has full knowledge of the architecture of the detector model, the algorithm that is used to create watermarks and the detector, hyperparameters used in the algorithm, and the exact training set used, and so represents a relatively strong attack.

\section{Conclusion}
\label{sec:conclusion}

We presented a general solution, that leverages imperceptible watermark, to the problem of detecting the provenance of a signal. In a departure from related work, our watermarking
schemes attempts to \emph{learn} constructions of watermarks that are resilient to adversarial transformations. Our solution can be applied in numerous environments such as in image, audio and video domains. Results presented both on images and audio suggest that it is possible to construct watermarks that simultaneously maintain a high level of signal fidelity, and are resilient to adversarial transformations, with a minimal
false positive rate.

Determining the provenance of a digital signal is becoming increasingly important. 
As machine learning methods advance, it is no longer clear if an image or segment of audio is authentic; plausible deniability has been eroded by these advances. 
In the coming years, machine learning techniques are likely to dominate traditional digital media processing pipelines, and so it is of paramount importance that the machine learning community develops techniques that can track and verify ownership of a signal. We view this work as a  step towards this ultimate goal.

\bibliographystyle{plainnat}
\bibliography{main}

\clearpage

\appendix
\onecolumn

\section{Evaluation on Cifar10 \& further evaluation on ImageNet and text-to-speech data}
\label{sec:eval_on_cifar10}

\begin{figure*}[htb!]
  \centering
    \begin{subfigure}[t]{0.3\textwidth}
        \centering
        \includegraphics[width=1.0\textwidth]{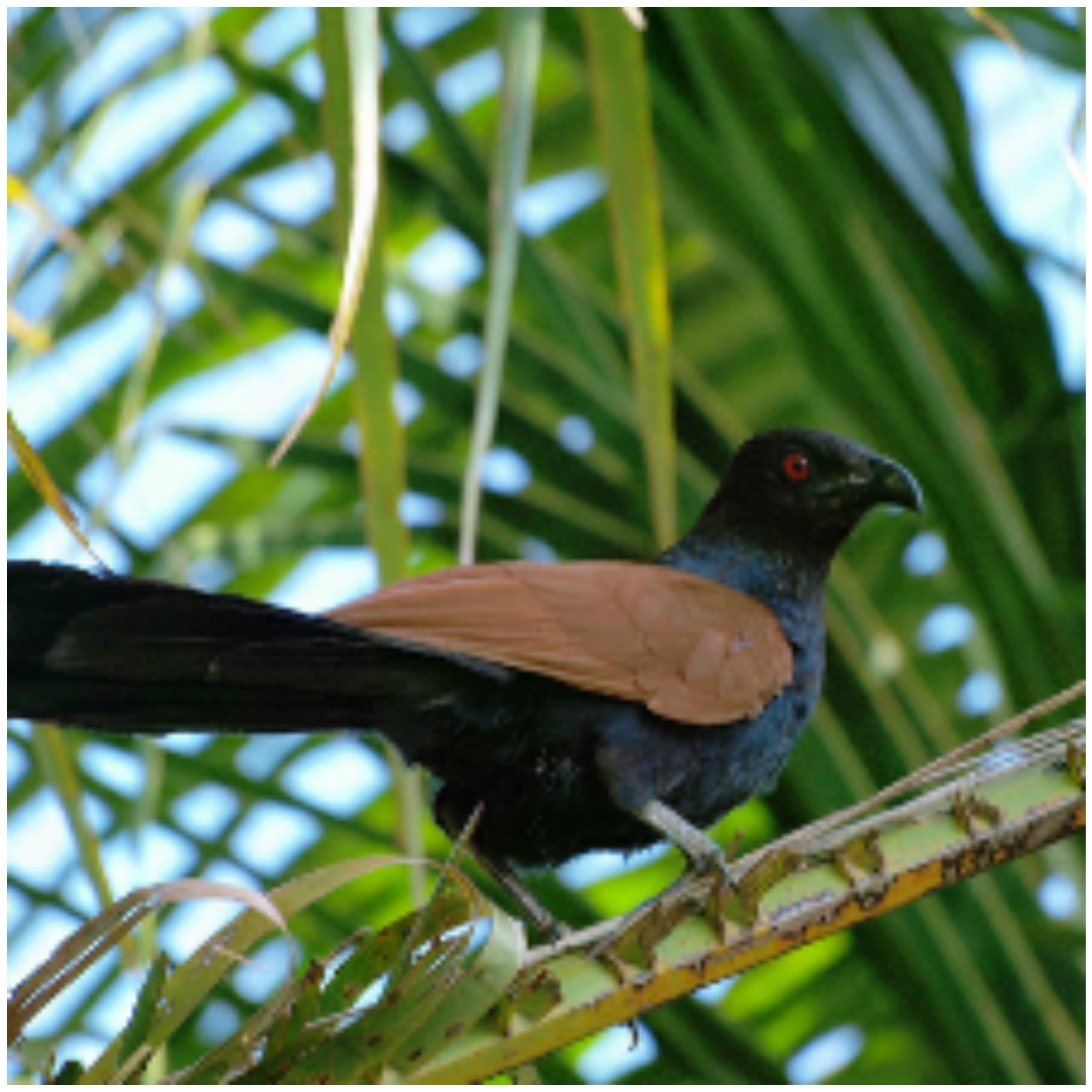}
        \label{fig:wm_ssim_0_98}
        \caption{Not watermarked}
    \end{subfigure}
    \begin{subfigure}[t]{0.3\textwidth}
        \centering
        \includegraphics[width=1.0\textwidth]{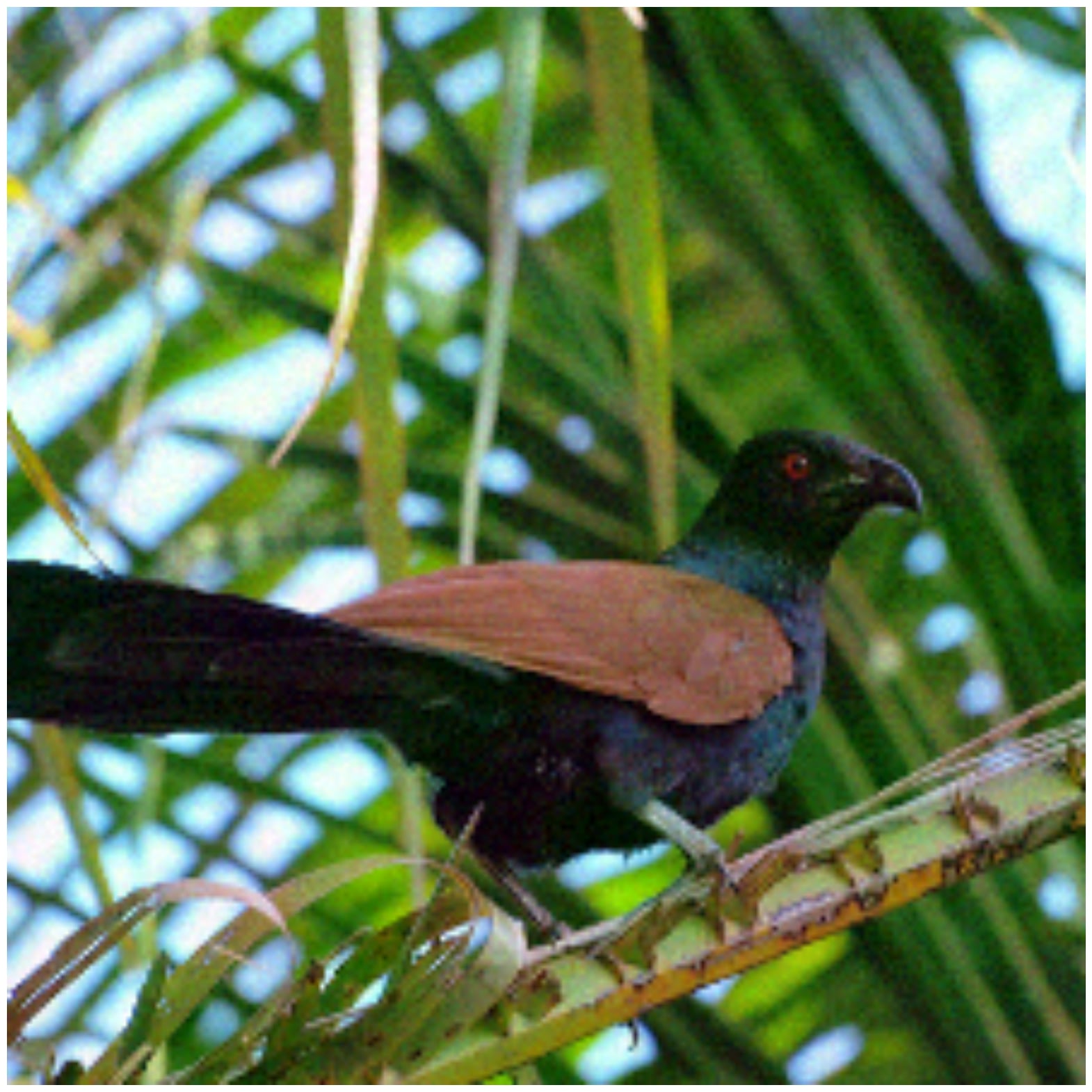}
        \label{fig:wm_ssim_0_82}
        \caption{Watermarked (SSIM 0.82)}
    \end{subfigure}
    \begin{subfigure}[t]{0.3\textwidth}
        \centering
        \includegraphics[width=1.0\textwidth]{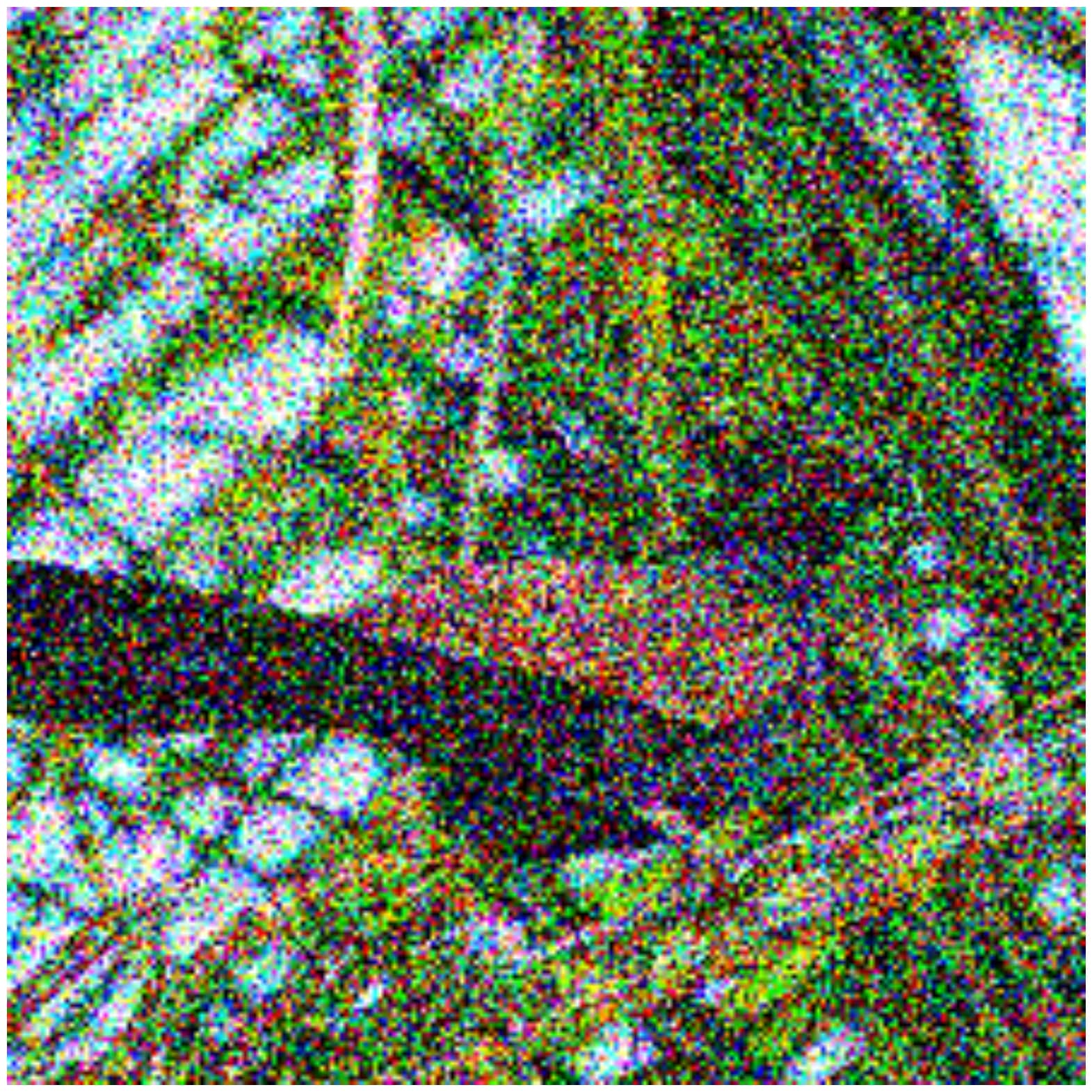}
        \label{fig:twm_ssim_0_1_sigma_0_4}
        \caption{Watermarked + Gaussian noise ($\sigma=0.4$, SSIM 0.1)}
    \end{subfigure}
    \caption{Sample of a transformation attack on ImageNet ($f_{\text{comp}}^{\epsilon_{10}}$). (a) is a non-watermarked test set image, (b) is a watermarked version with the minimum necessary watermark required to be correctly classified by a detection model when Gaussian noise is added at a standard deviation depicted in (c).}
   \label{fig:imagenet_full_acc_noise_examples}
\end{figure*}

\begin{figure*}[htb!]
    \centering
    \begin{subfigure}[t]{0.24\textwidth}
        \centering
        \includegraphics[width=1.0\textwidth]{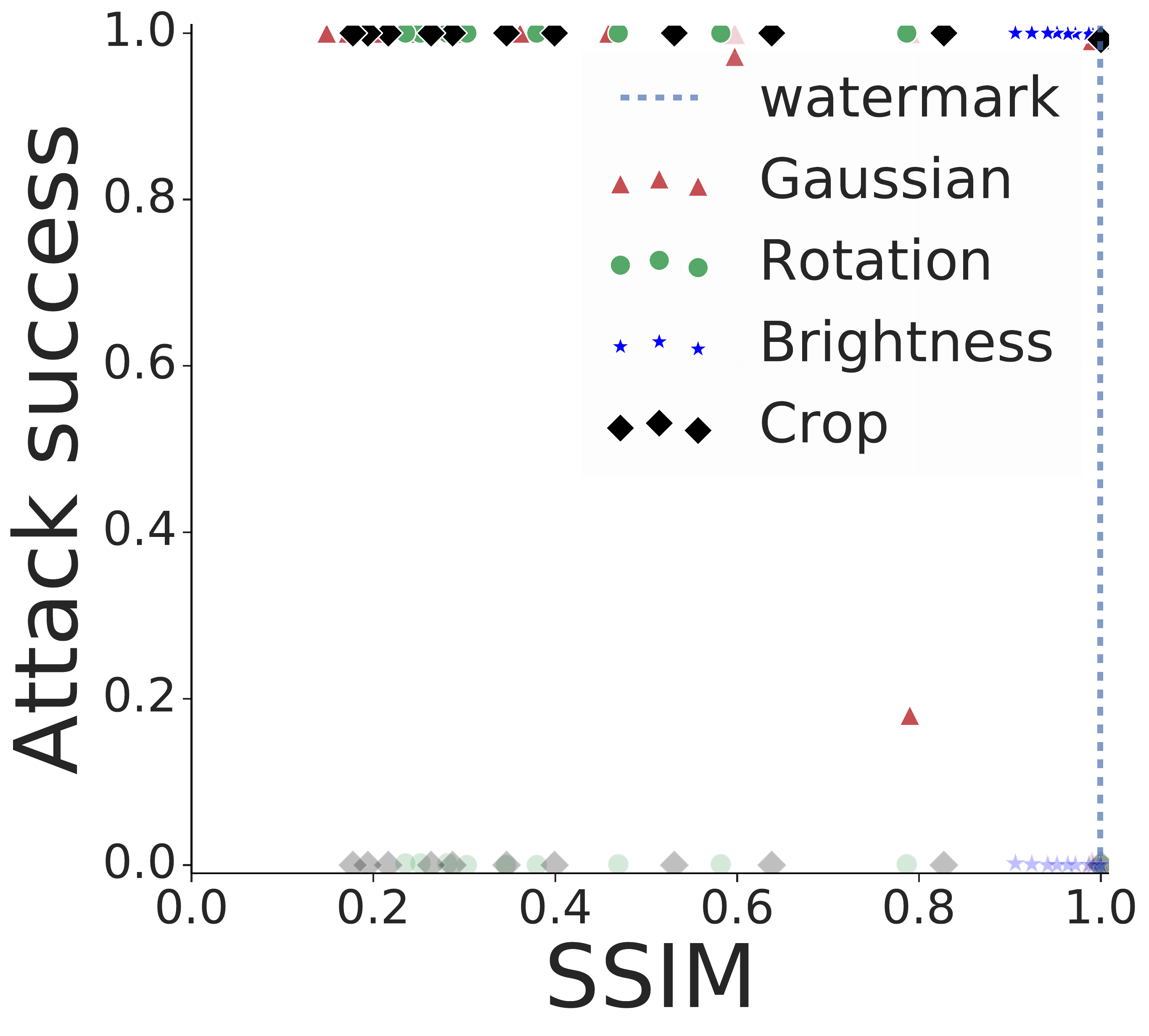}
        \label{fig:cifar_10_rta_eps_1_f1}
        \caption{$f_{\text{{\tiny CIFAR}}}^1$, $\epsilon=\nicefrac{1}{255}$}
    \end{subfigure}
    \begin{subfigure}[t]{0.24\textwidth}
        \centering
        \includegraphics[width=1.0\textwidth]{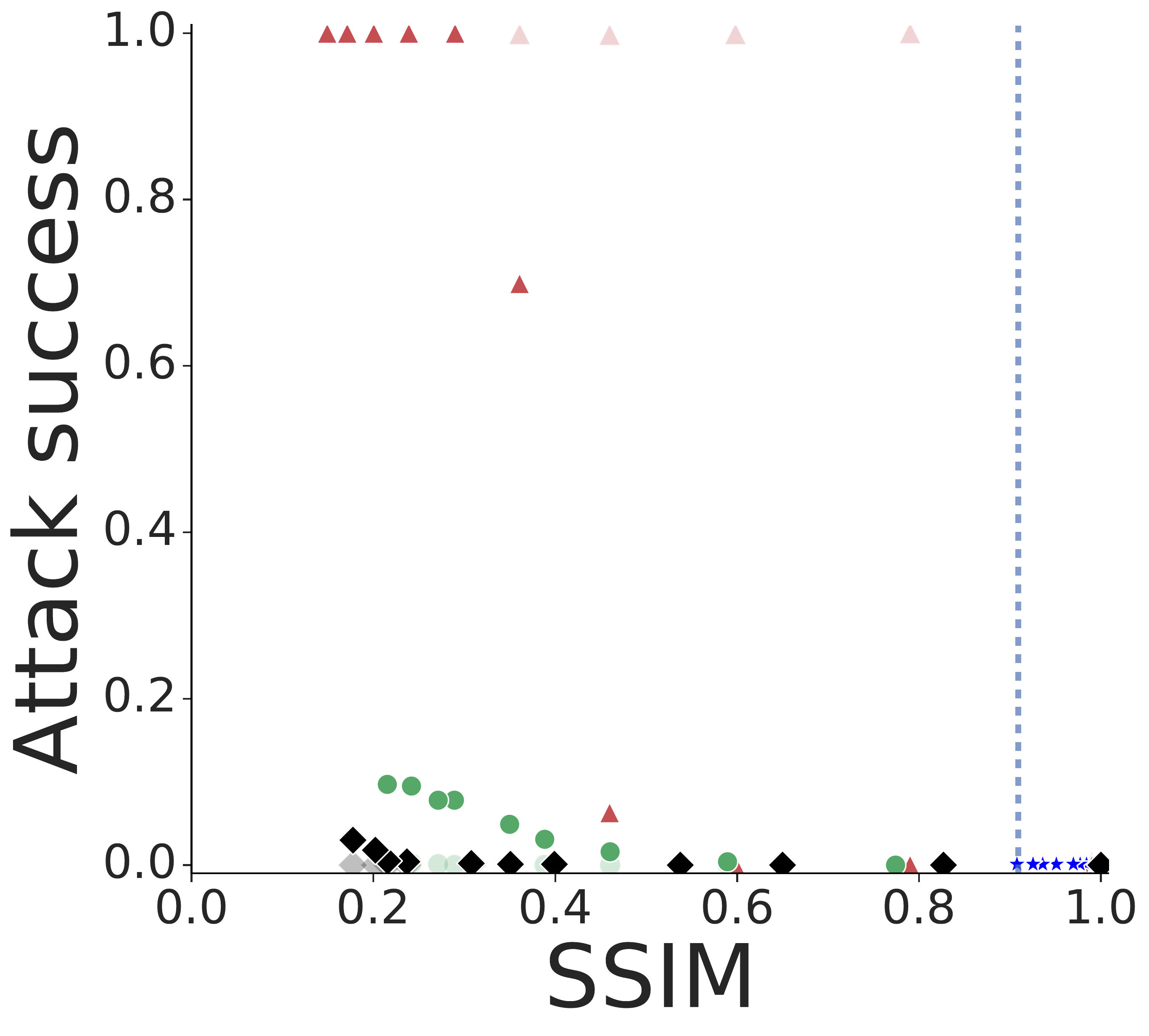}
        \label{fig:cifar_10_rta_eps_12_f1}
        \caption{$f_{\text{{\tiny CIFAR}}}^1$, $\epsilon=\nicefrac{12}{255}$}
    \end{subfigure}
    \begin{subfigure}[t]{0.24\textwidth}
        \centering
        \includegraphics[width=1.0\textwidth]{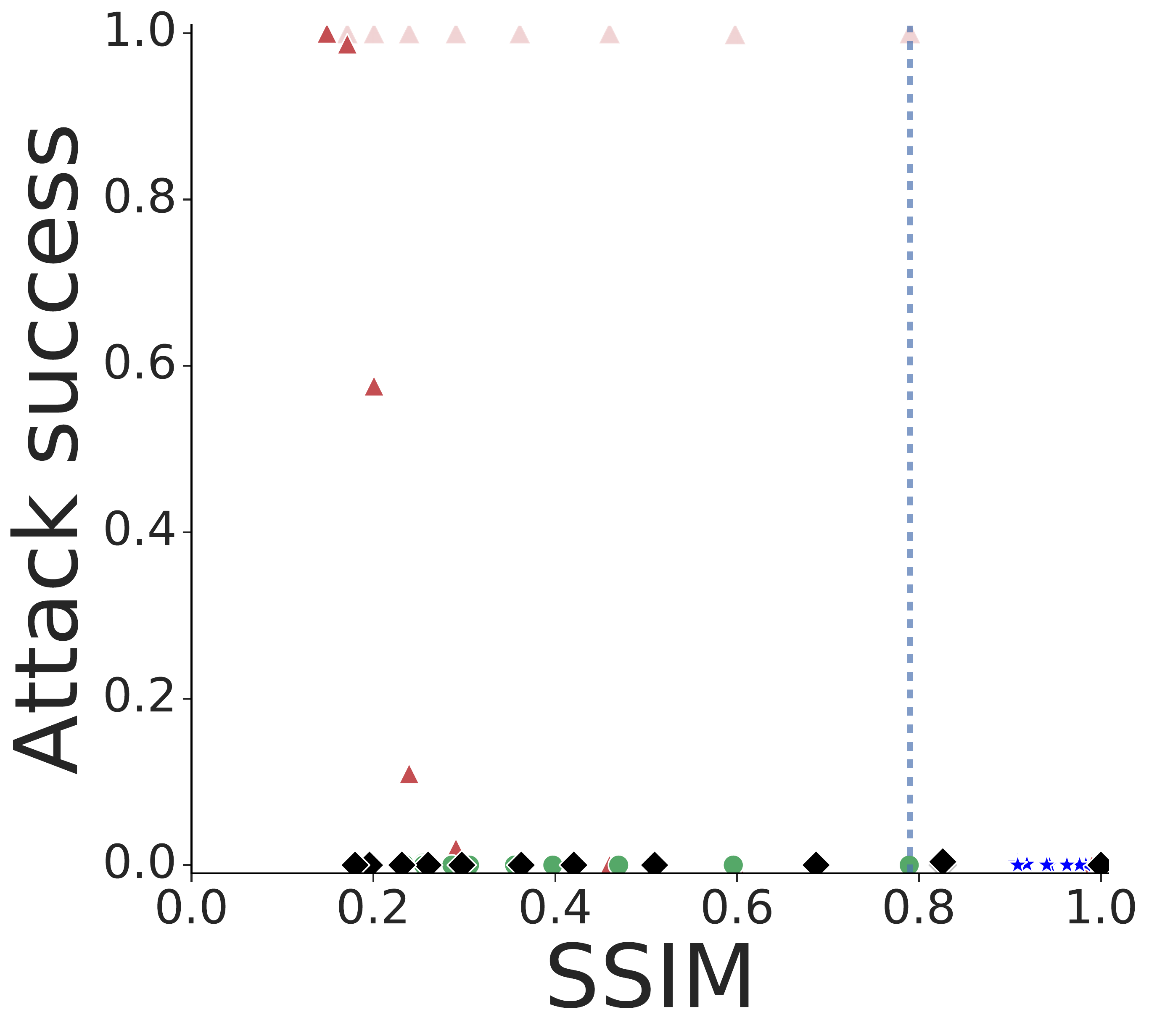}
        \label{fig:cifar_10_rta_eps_23_f1}
        \caption{$f_{\text{{\tiny CIFAR}}}^1$, $\epsilon=\nicefrac{23}{255}$}
    \end{subfigure}
    \begin{subfigure}[t]{0.24\textwidth}
        \centering
        \includegraphics[width=1.0\textwidth]{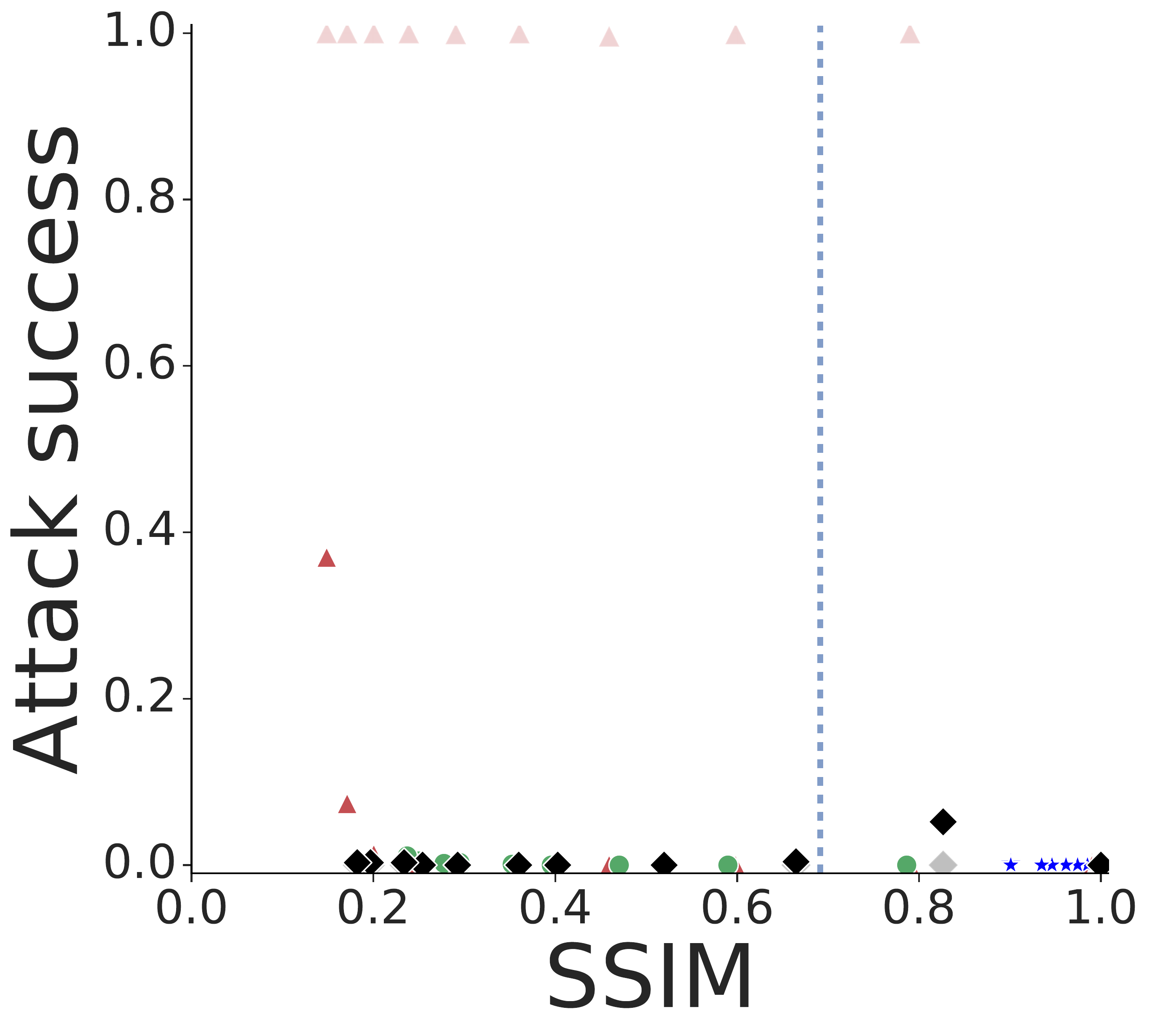}
        \label{fig:cifar_10_rta_eps_33_f1}
        \caption{$f_{\text{{\tiny CIFAR}}}^1$, $\epsilon=\nicefrac{33}{255}$}
    \end{subfigure}
    \vskip\baselineskip
    \begin{subfigure}[t]{0.24\textwidth}
        \centering
        \includegraphics[width=1.0\textwidth]{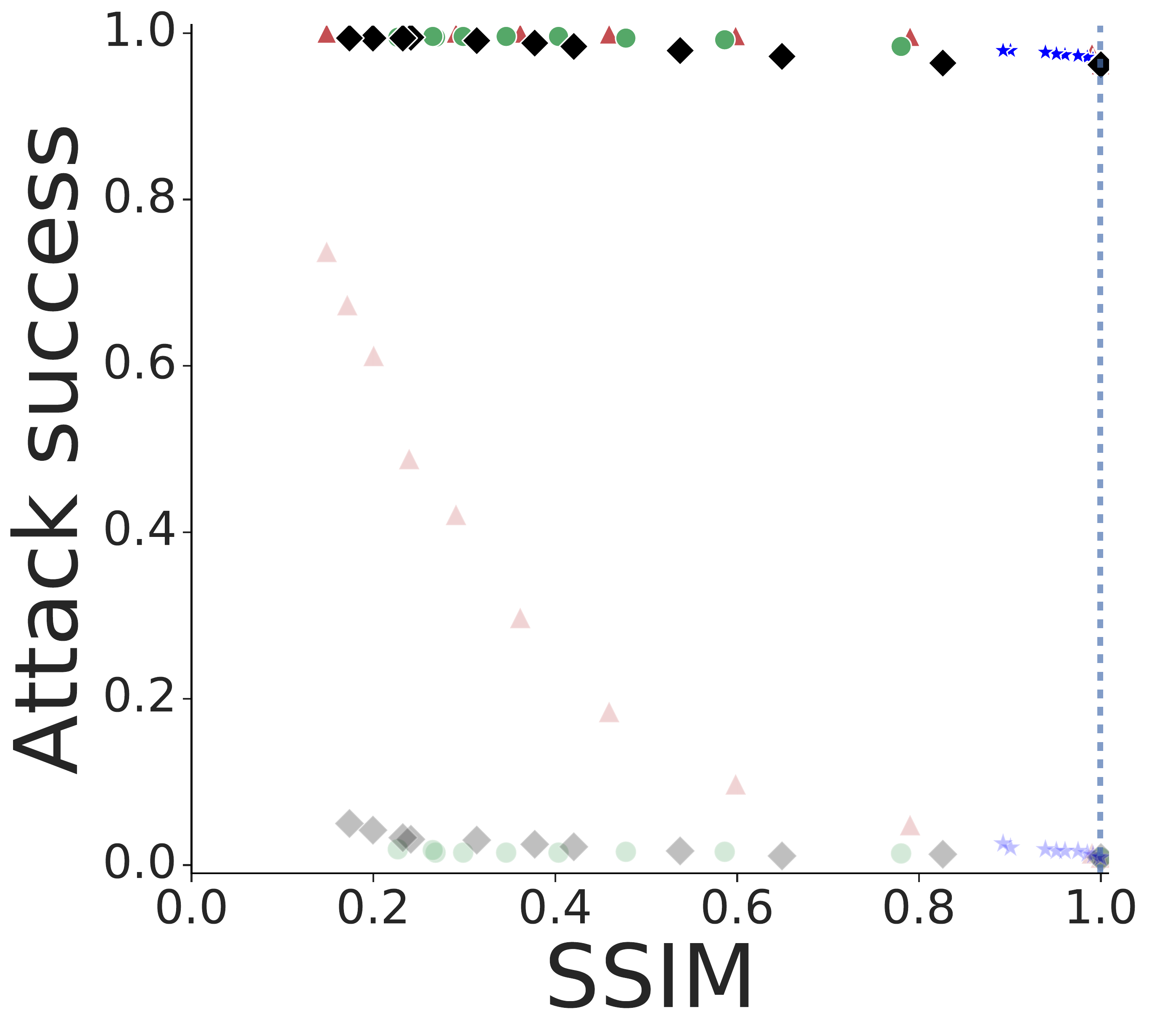}
        \label{fig:cifar_10_rta_eps_1_f2}
        \caption{$f_{\text{{\tiny CIFAR}}}^2$, $\epsilon=\nicefrac{1}{255}$}
    \end{subfigure}
    \begin{subfigure}[t]{0.24\textwidth}
        \centering
        \includegraphics[width=1.0\textwidth]{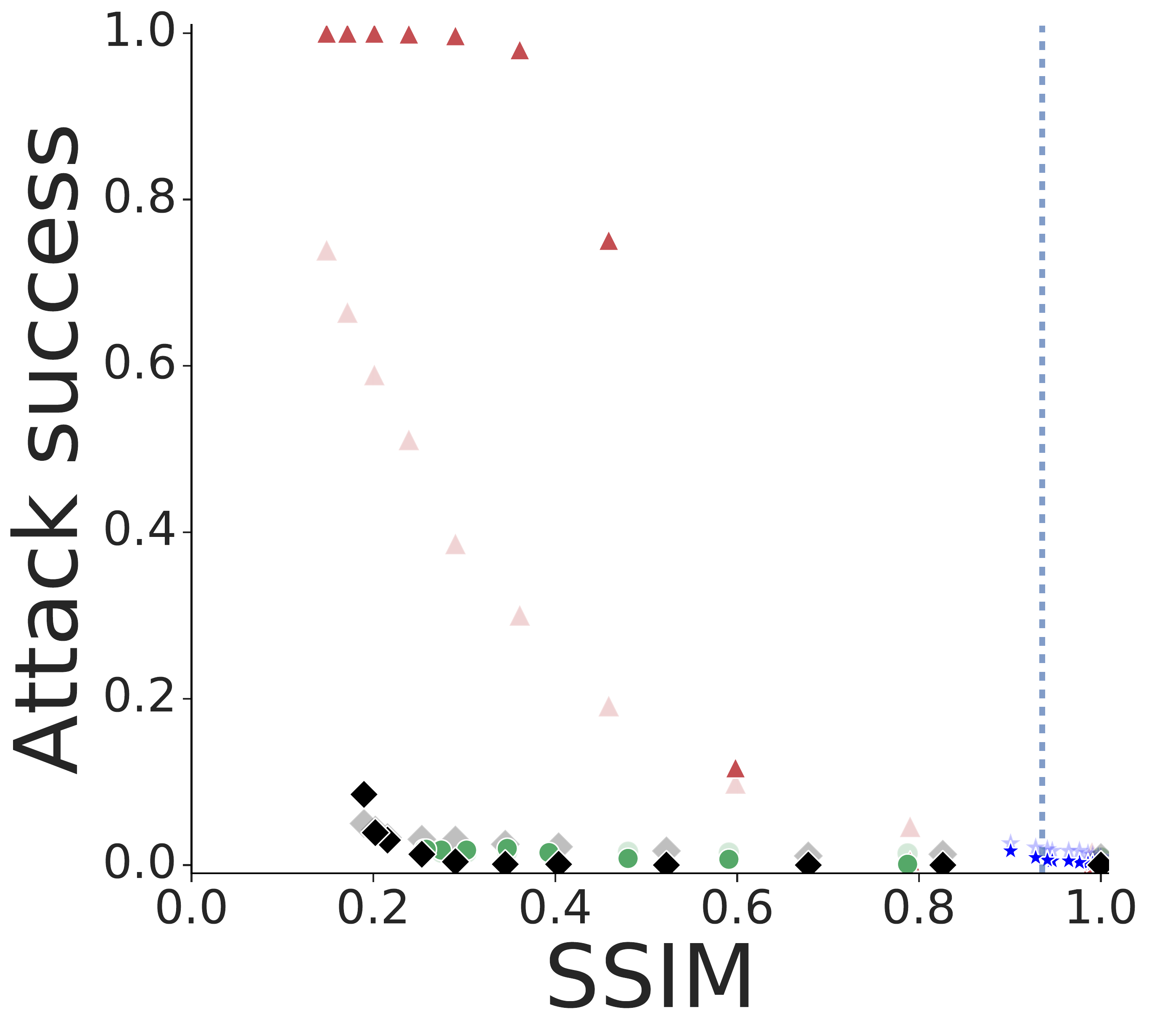}
        \label{fig:cifar_10_rta_eps_12_f2}
        \caption{$f_{\text{{\tiny CIFAR}}}^2$, $\epsilon=\nicefrac{12}{255}$}
    \end{subfigure}
    \begin{subfigure}[t]{0.24\textwidth}
        \centering
        \includegraphics[width=1.0\textwidth]{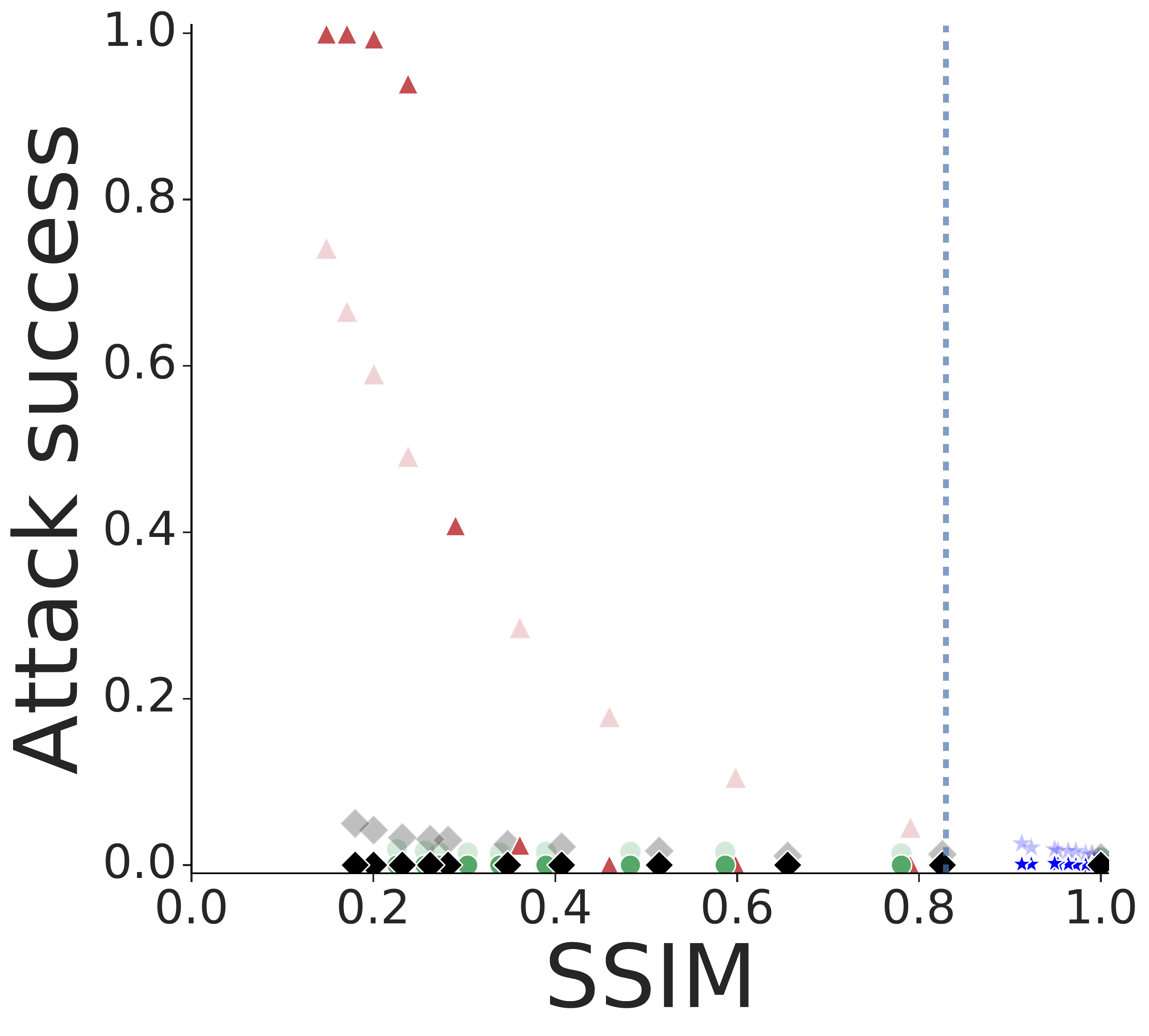}
        \label{fig:cifar_10_rta_eps_23_f2}
        \caption{$f_{\text{{\tiny CIFAR}}}^2$, $\epsilon=\nicefrac{23}{255}$}
    \end{subfigure}
    \begin{subfigure}[t]{0.24\textwidth}
        \centering
        \includegraphics[width=1.0\textwidth]{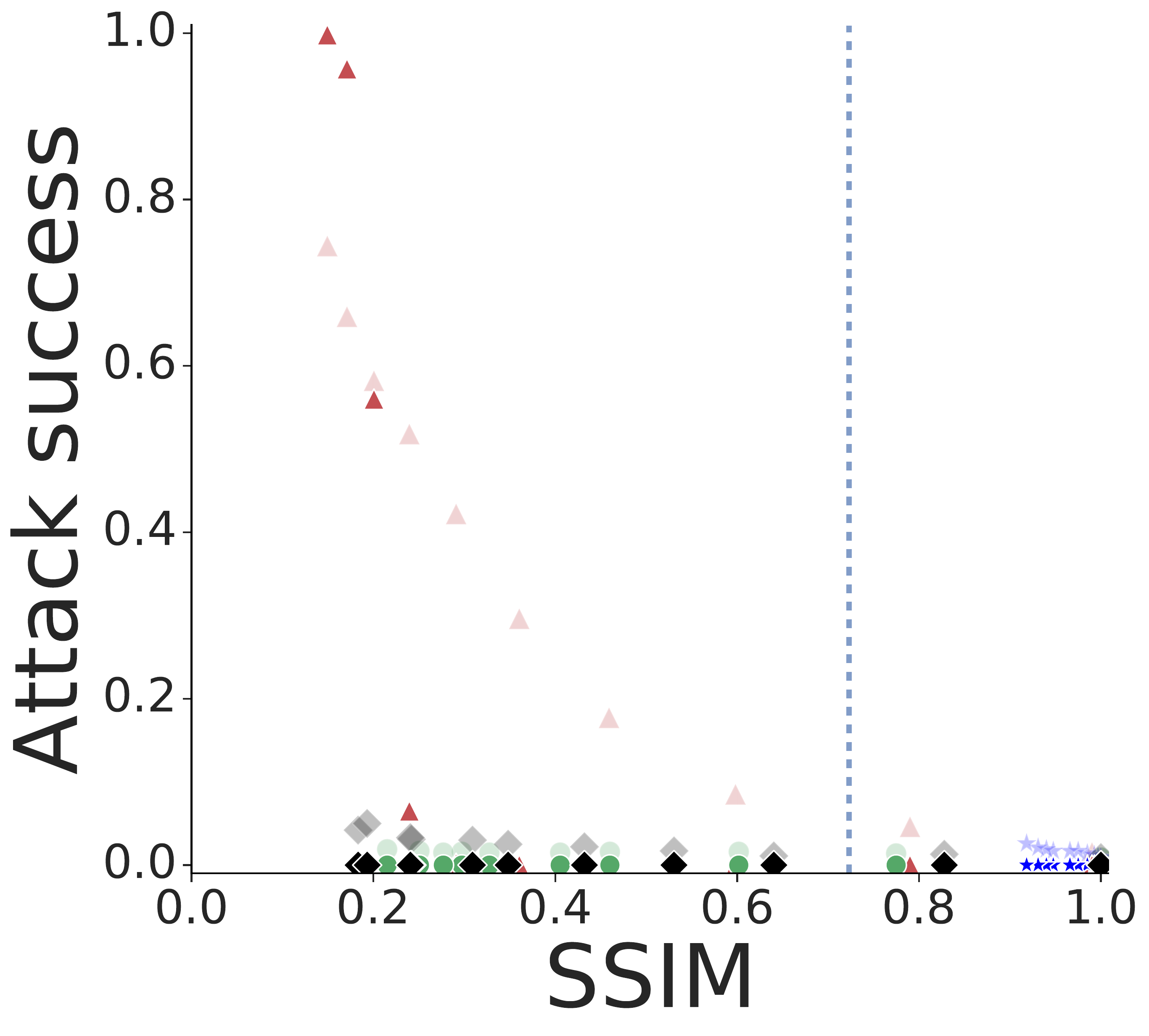}
        \label{fig:cifar_10_rta_eps_33_f2}
        \caption{$f_{\text{{\tiny CIFAR}}}^2$, $\epsilon=\nicefrac{33}{255}$}
    \end{subfigure}
    \caption{Transformation attack results on Cifar10. Each subfigure is plotted as function
  of the distortion introduced by the transformation, and we evaluate for a watermark perturbation ranging
  from $\epsilon=\nicefrac{1}{255}$ to $\epsilon=\nicefrac{33}{255}$. The solid markers are the attack success
  against watermarked images, and the faded markers, the attack success on
  non-watermarked images.}
   \label{fig:cifar10_transformation_attack_fine}
\end{figure*}

\noindent \textbf{Cifar10 \citep{cifar10}.} We use a wide ResNet classifier \citep{he2016deep} for the watermark detector. We replace batch normalization with instance
normalization \citep{ulyanov2016instance}.
We trained with a mini-batch size of 32 for 60,000 steps, with an initial learning rate of 0.01 and decaying this by a factor of 10 every 20,000 steps. All images are normalized into the range [0,1]. We set the maximum size of the watermark perturbation, 
$\epsilon$, to $\nicefrac{20}{255}$, and the watermark perturbation is constructed with five PGD steps at each iteration during training. With respect to the set of transformations, we set $\sigma$ to 0.5,
$r$ to $\nicefrac{\pi}{2}$, $c_h$ and $c_w$ to 2, and $b$ to 0.25. We trained two classifiers under these 
hyperparameters, one where at each step we randomly sample a single transformation, and another classifier
where we apply a composition of all transformations, which shall be denoted by $f_{\text{{\tiny CIFAR}}}^1$ and 
$f_{\text{{\tiny CIFAR}}}^2$, respectively. Both classifiers achieved $100\%$ test set accuracy, where the test set consists of 10,000 non-watermarked images and 10,000 watermarked
images.

\begin{figure*}[t]
    \centering
    \begin{subfigure}[t]{0.24\textwidth}
        \centering
        \includegraphics[width=1.0\textwidth]{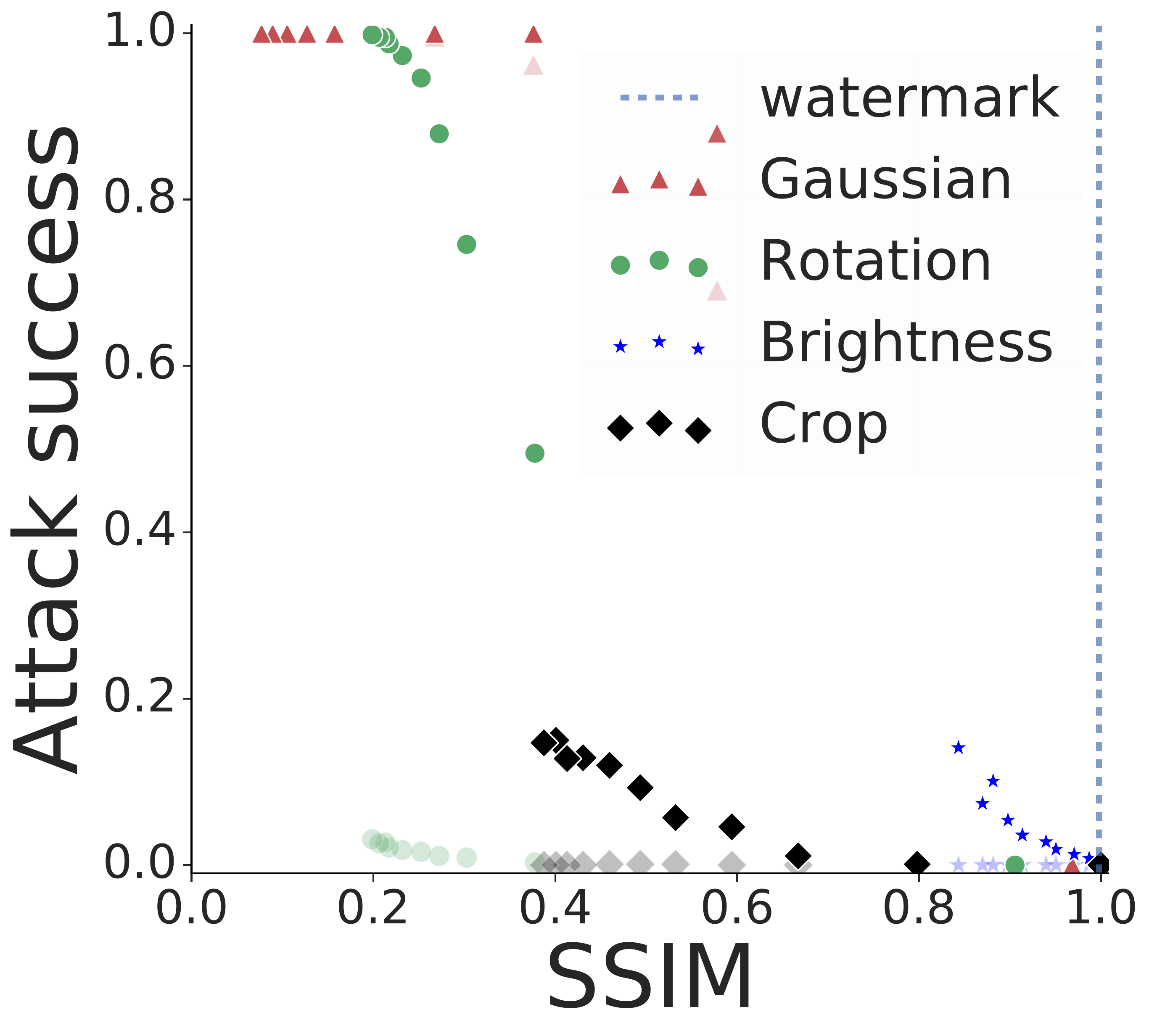}
        \label{fig:imagenet_rta_eps_1_wid_1}
        \caption{$f^{\epsilon_{5}}$, $\epsilon=\nicefrac{1}{255}$}
    \end{subfigure}
    \begin{subfigure}[t]{0.24\textwidth}
        \centering
        \includegraphics[width=1.0\textwidth]{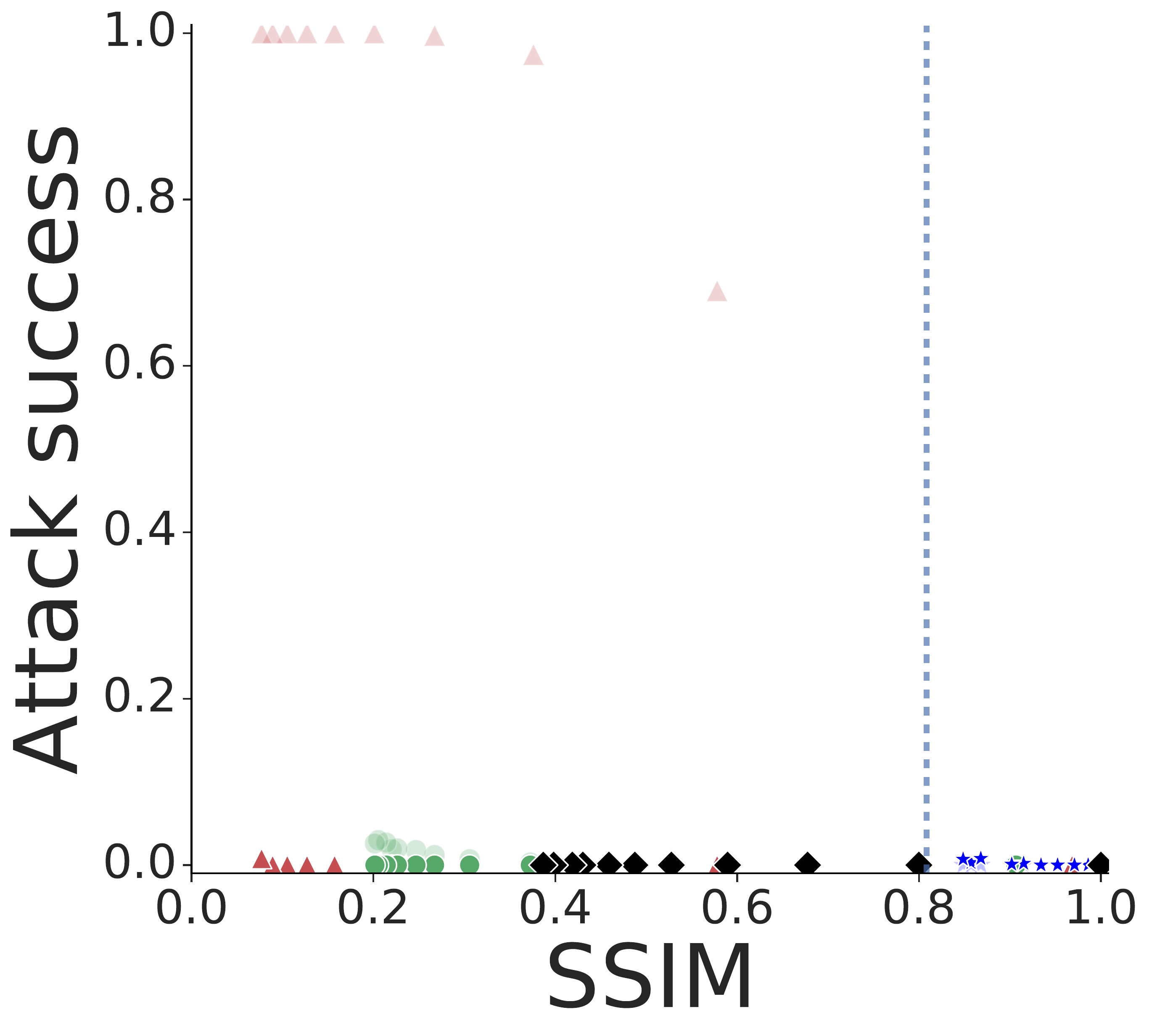}
        \label{fig:imagenet_rta_eps_12_wid_1}
        \caption{$f^{\epsilon_{5}}$, $\epsilon=\nicefrac{12}{255}$}
    \end{subfigure}
    \begin{subfigure}[t]{0.24\textwidth}
        \centering
        \includegraphics[width=1.0\textwidth]{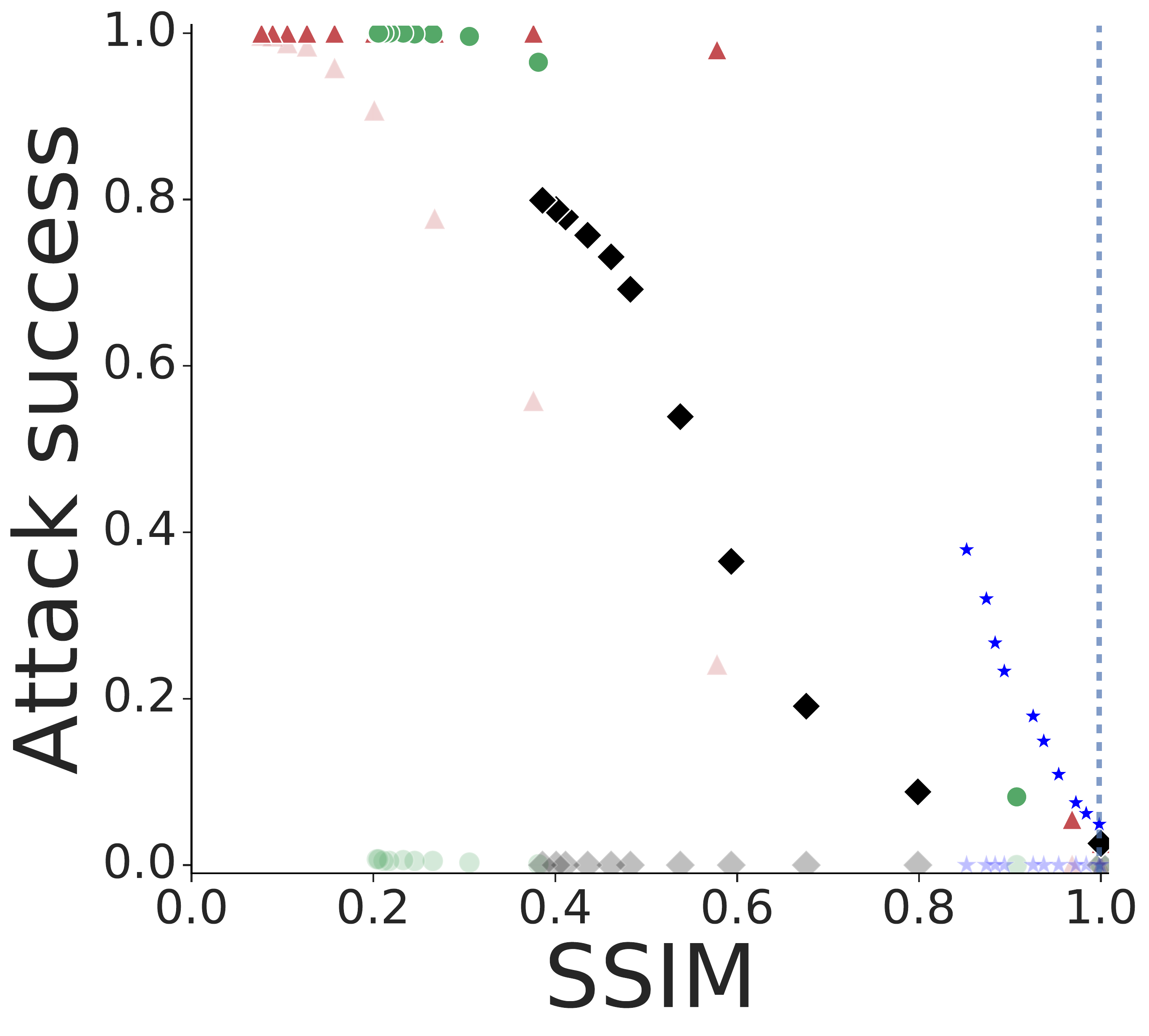}
        \label{fig:imagenet_rta_eps_1_wid_2}
        \caption{$f^{\epsilon_{10}}$, $\epsilon=\nicefrac{1}{255}$}
    \end{subfigure}
    \begin{subfigure}[t]{0.24\textwidth}
        \centering
        \includegraphics[width=1.0\textwidth]{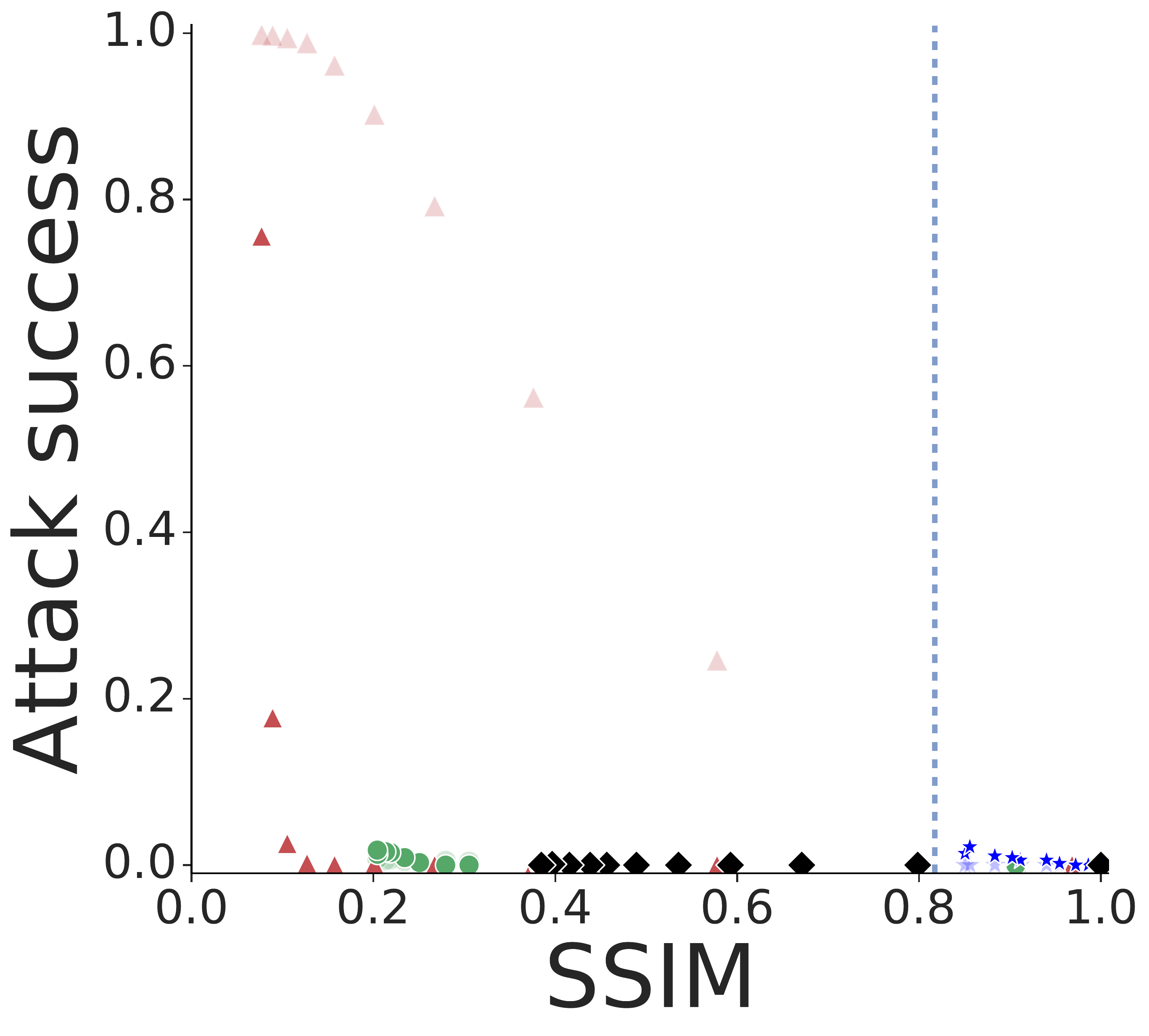}
        \label{fig:imagenet_rta_eps_12_wid_2}
        \caption{$f^{\epsilon_{10}}$, $\epsilon=\nicefrac{12}{255}$}
    \end{subfigure}
    \vskip\baselineskip
    \begin{subfigure}[t]{0.24\textwidth}
        \centering
        \includegraphics[width=1.0\textwidth]{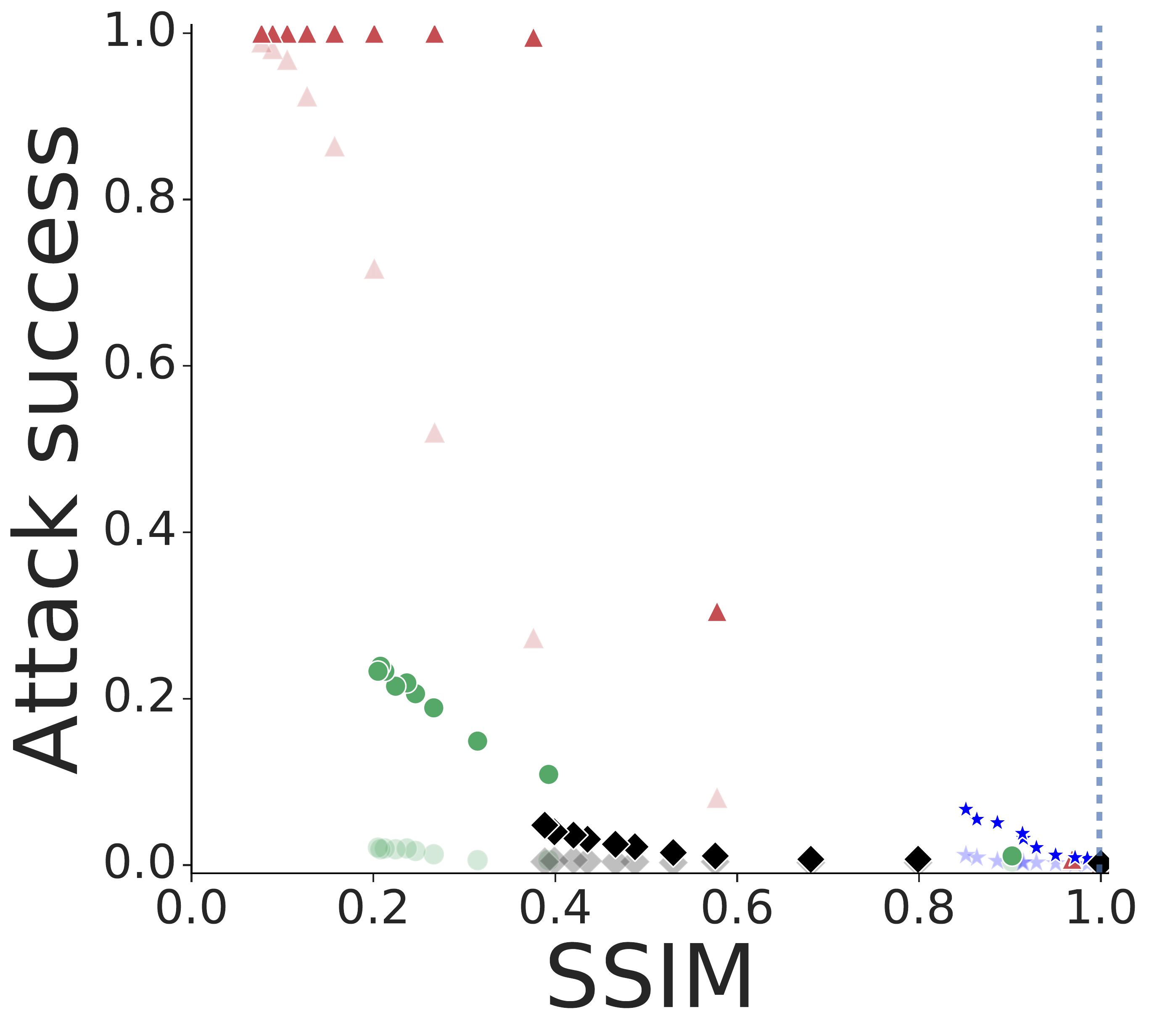}
        \label{fig:imagenet_rta_eps_1_wid_3}
        \caption{$f_{\text{comp}}^{\epsilon_{5}}$, $\epsilon=\nicefrac{1}{255}$}
    \end{subfigure}
    \begin{subfigure}[t]{0.24\textwidth}
        \centering
        \includegraphics[width=1.0\textwidth]{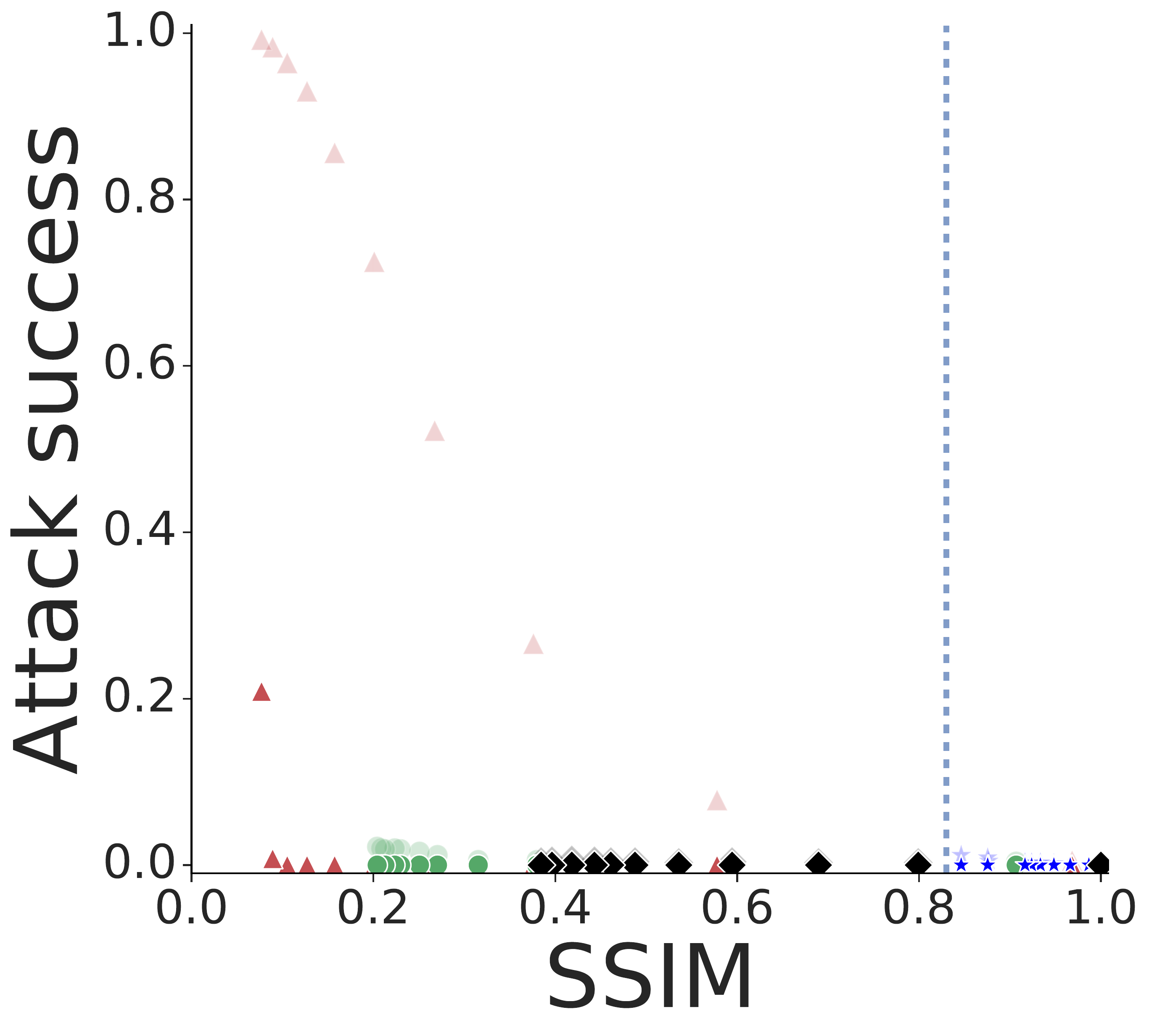}
        \label{fig:imagenet_rta_eps_12_wid_3}
        \caption{$f_{\text{comp}}^{\epsilon_{5}}$, $\epsilon=\nicefrac{12}{255}$}
    \end{subfigure}
    \begin{subfigure}[t]{0.24\textwidth}
        \centering
        \includegraphics[width=1.0\textwidth]{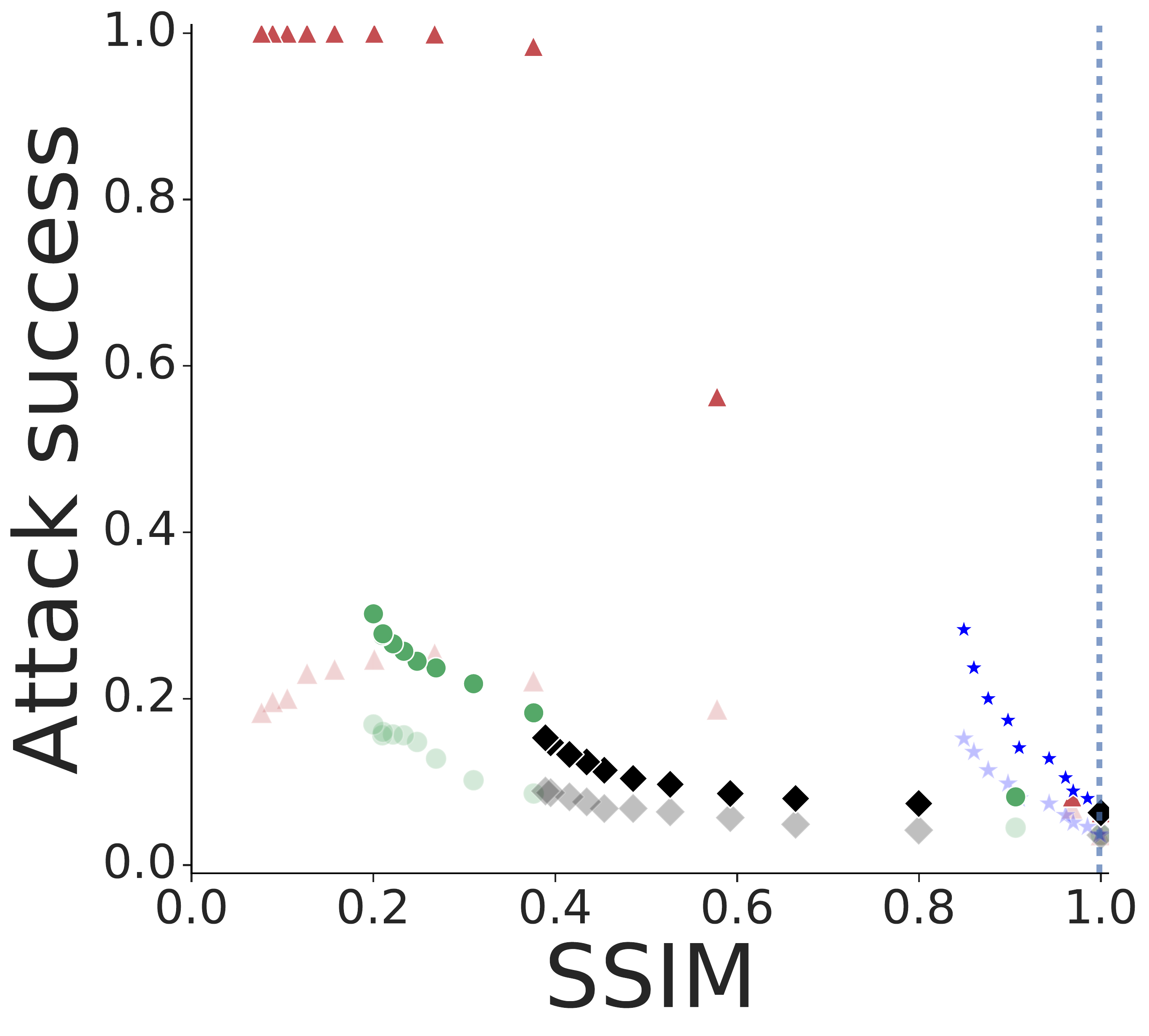}
        \label{fig:imagenet_rta_eps_1_wid_4}
        \caption{$f_{{{\text{comp}}}}^{\epsilon_{10}}$, $\epsilon=\nicefrac{1}{255}$}
    \end{subfigure}
    \begin{subfigure}[t]{0.24\textwidth}
        \centering
        \includegraphics[width=1.0\textwidth]{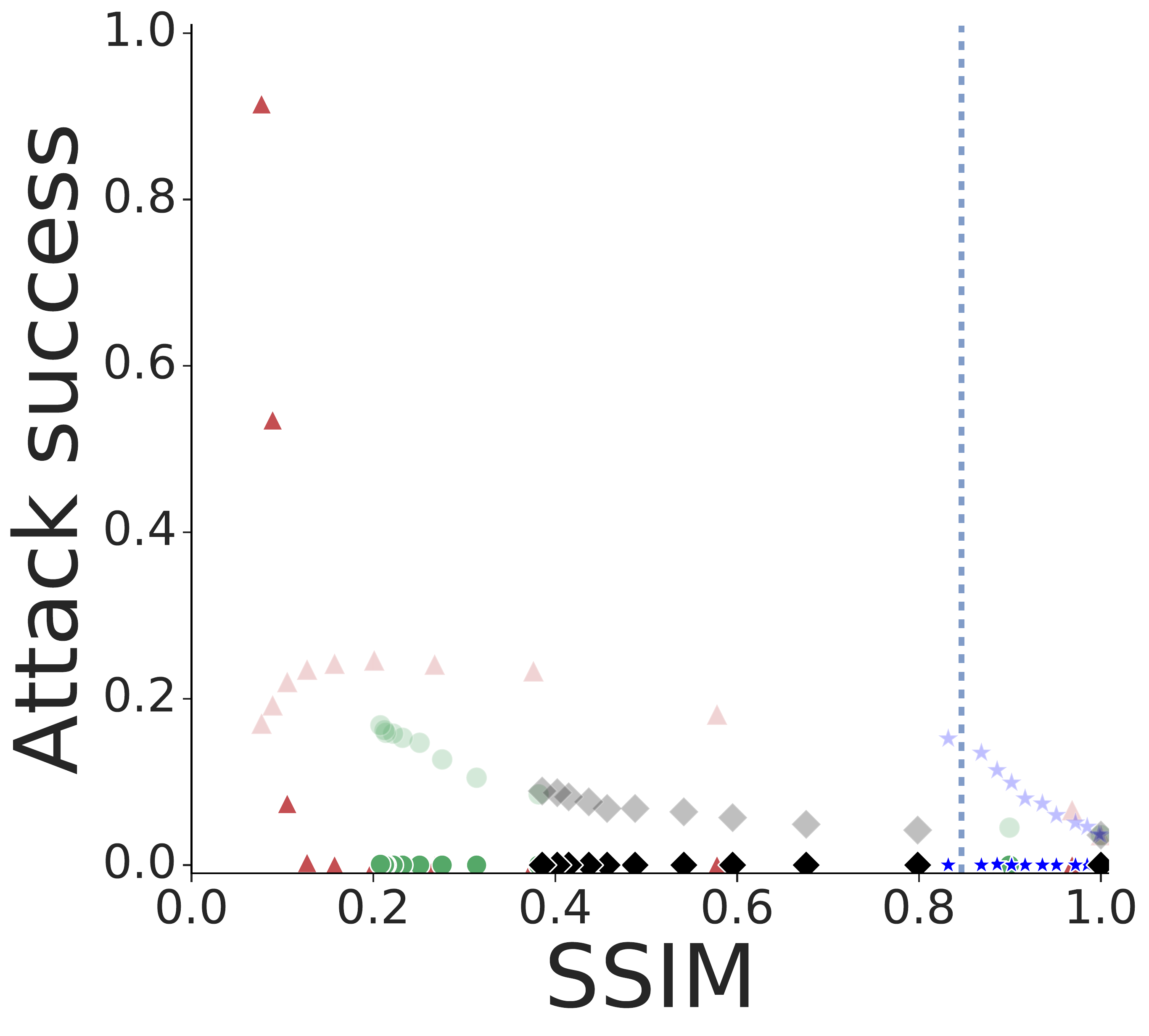}
        \label{fig:imagenet_rta_eps_12_wid_4}
        \caption{$f_{{{\text{comp}}}}^{\epsilon_{10}}$, $\epsilon=\nicefrac{12}{255}$}
    \end{subfigure}
    \caption{Transformation attack results on ImageNet. Each subfigure is plotted as function
  of the distortion introduced by the transformation, and we evaluate for a watermark perturbation ranging
  from $\epsilon=\nicefrac{1}{255}$ to $\epsilon=\nicefrac{12}{255}$. The solid markers are the attack success
  against watermarked images, and the faded markers, the attack success on
  non-watermarked images.}
   \label{fig:imagenet_transformation_attack_fine}
\end{figure*}

\begin{figure*}[t!]
    \centering
    \begin{subfigure}[t]{0.4\textwidth}
        \centering
        \includegraphics[width=1.0\textwidth]{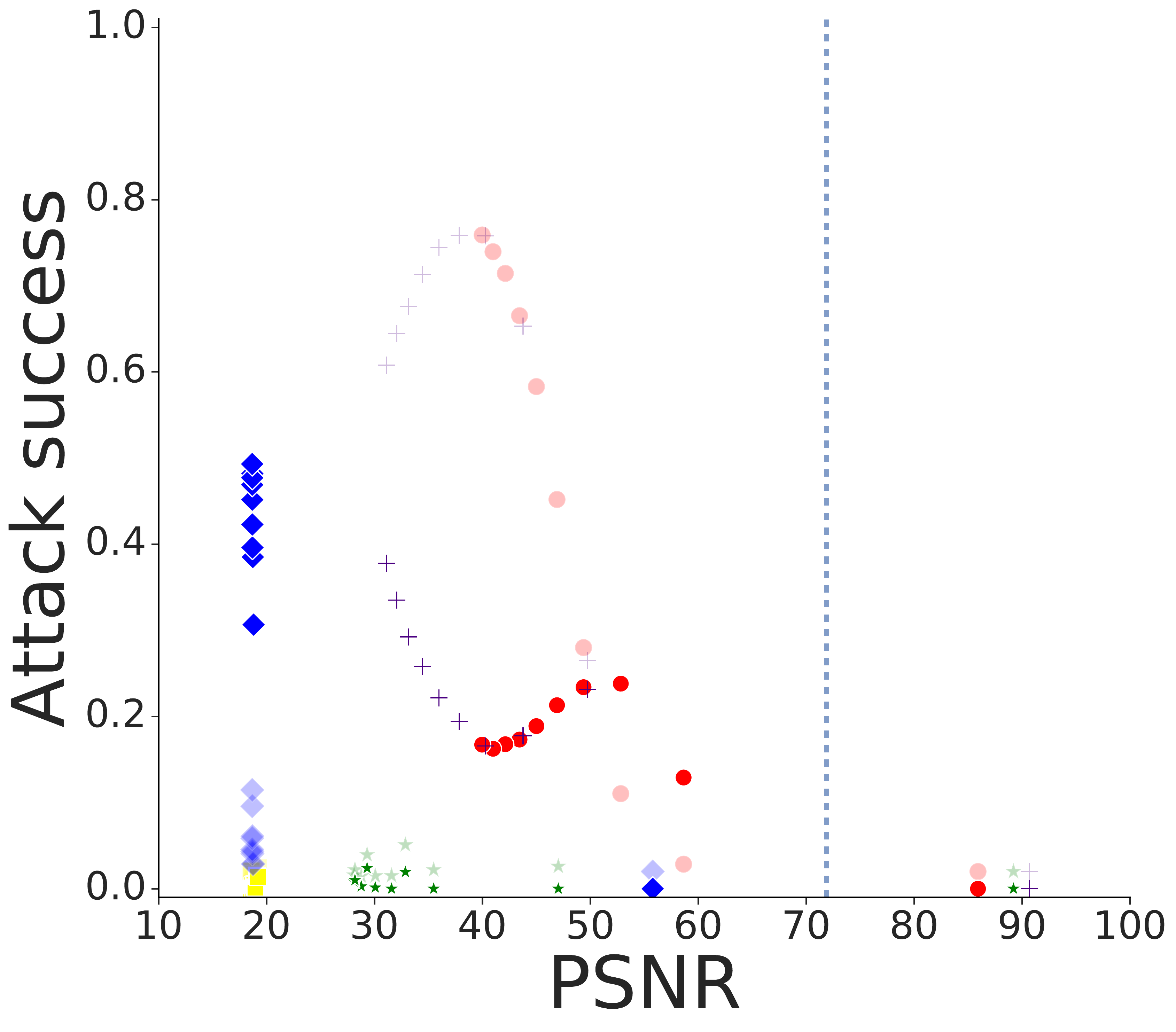}
        \label{fig:tts_rta_wx_eps_small}
        \caption{Watermarked with $\epsilon=4\times10^{-4}$}
    \end{subfigure}
    \begin{subfigure}[t]{0.4\textwidth}
        \centering
        \includegraphics[width=1.0\textwidth]{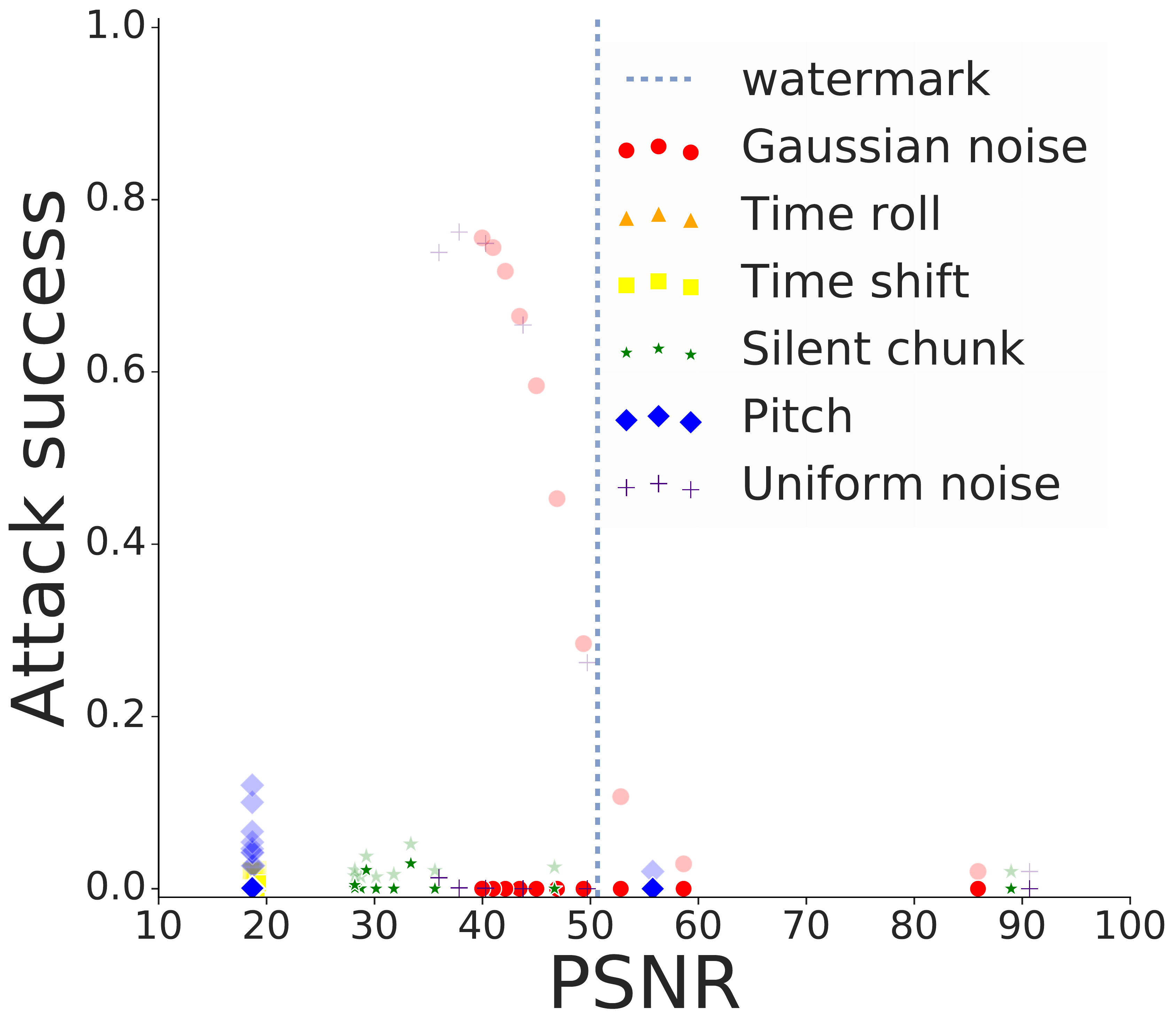}
        \label{fig:tts_rta_wx_eps_big}
        \caption{Watermarked with $\epsilon=4.8\times10^{-3}$}
    \end{subfigure}
    \caption{Transformation attack results on audio data. Each marker represents the average attack success
    versus average PSNR of 10,000 inputs sampled under a transformation.  The solid markers are the attack success
  against watermarked images, and the faded markers, the attack success on
  non-watermarked images.}
   \label{fig:tts_rta_fine}
\end{figure*}

Given a transformation function $t_{\theta}:\mathcal{X} \rightarrow \mathcal{X}$, parameterized by $\theta$ which encodes some randomness, and an input $s\in\mathcal{X}$ with class
$y\in\{0,1\}$, we create $n$ copies of this input under $t_{\theta}$, $s^i$, $i\in\{1,..., n\}$. We then measure both the SSIM of the watermarked images and
SSIM of both transformed watermarked and transformed non-watermarked images as a function of the attack success over these $n$ inputs. \Cref{fig:cifar10_transformation_attack_fine} shows the results of a transformation attack: for each of 2,000 Cifar10 test set images (1,000 non-watermarked and 1,000 watermarked)
we create 1000 new images under a transformation; each marker in the figure is the average attack success over 1 million test images. We evaluate the
attack success for a number of different watermarking $\epsilon$ values. Large $\epsilon$ increase the perceptibility of the watermark, and so correspondingly
decrease the SSIM score. Note, attacking non-watermarked content is unaffected by the $\epsilon$ value chosen for watermarking.

For $\epsilon = \nicefrac{1}{255}$, the SSIM score of watermarked content is $\approx0.99$, and thus the watermark is highly imperceptible. 
However, a number of attacks on watermark content succeed with high SSIM score, such as for example transformations
that modify the brightness of an image. Slightly increasing the perceptibly of the watermark to $\epsilon = \nicefrac{12}{255}$,
reduces the attack success of nearly all transformations to zero. Overall, the classifier trained
on a composition of transformations is more robust to attacks on non-watermarked content. 

Similar effects can be observed
on ImageNet shown in \cref{fig:imagenet_transformation_attack_fine}, and the propriety audio dataset in \cref{fig:tts_rta_fine}. Interestingly,
watermark classifiers on ImageNet seem to be more robust at smaller $\epsilon$
values than on Cifar10, we conjecture this is because there is a larger space in which to distribute
the watermark.

\section{Further Comparison with Broken Arrows}
\label{sec:imagenet_full_acc_complete_psnr}

\begin{figure*}[htb!]
    \centering
    \begin{subfigure}[t]{0.3\textwidth}
        \centering
        \includegraphics[width=1.0\textwidth]{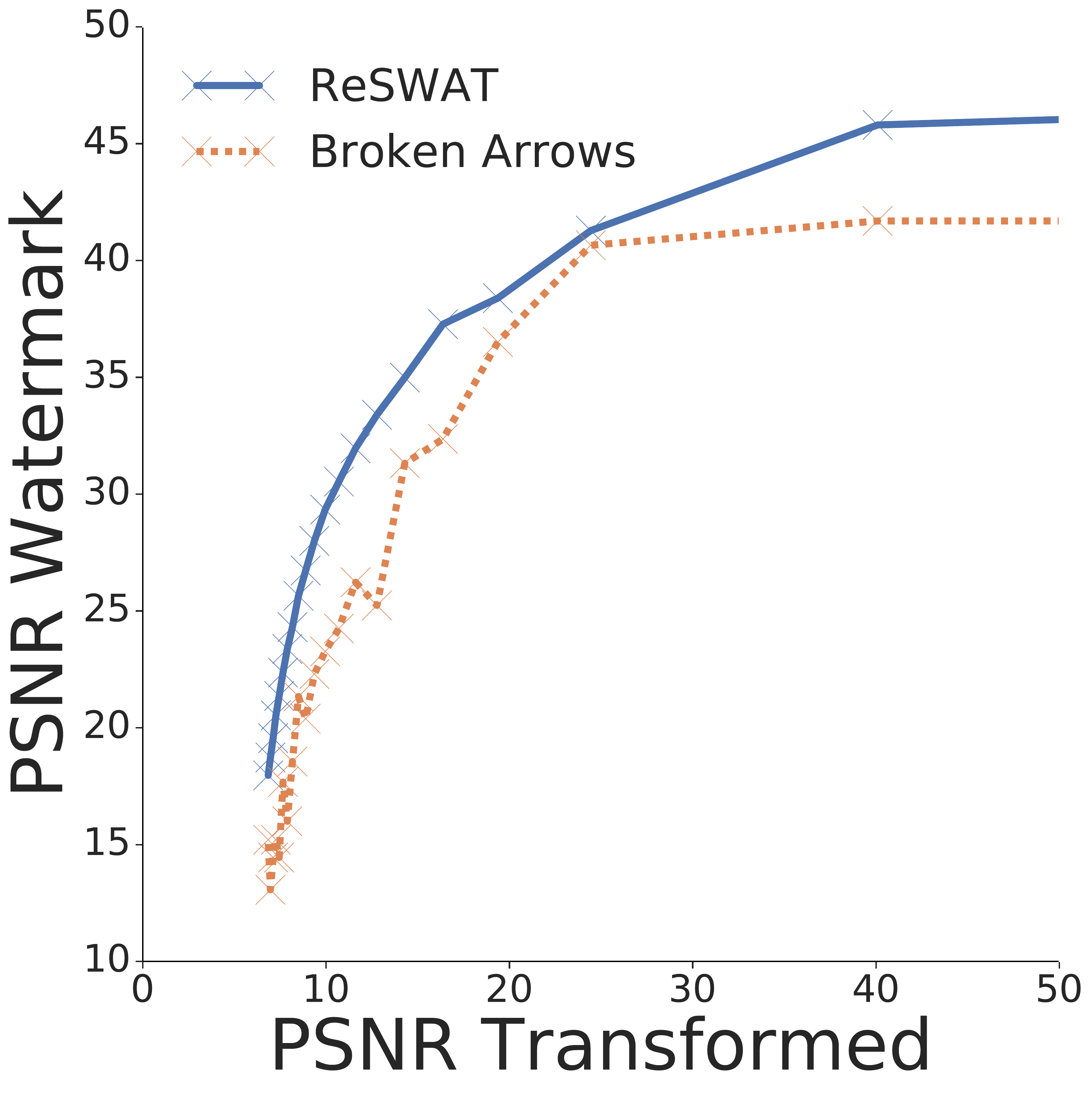}
        \label{fig:imagenet_full_acc_psnr_gaussian}
        \caption{Gaussian noise.}
    \end{subfigure}
    \begin{subfigure}[t]{0.3\textwidth}
        \centering
        \includegraphics[width=1.0\textwidth]{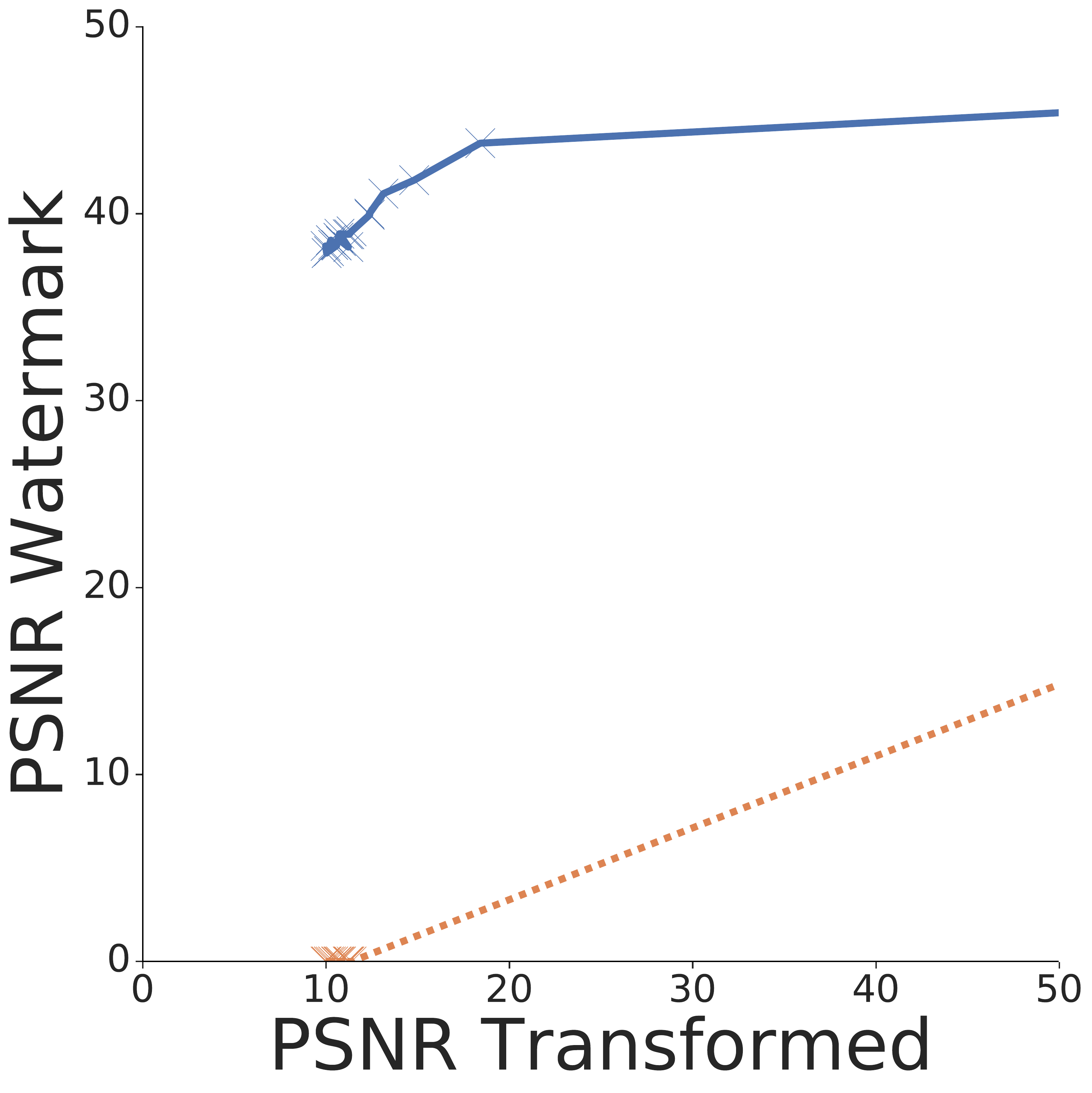}
        \label{fig:imagenet_full_acc_psnr_rotation}
        \caption{Rotation.}
    \end{subfigure}
    \begin{subfigure}[t]{0.3\textwidth}
        \centering
        \includegraphics[width=1.0\textwidth]{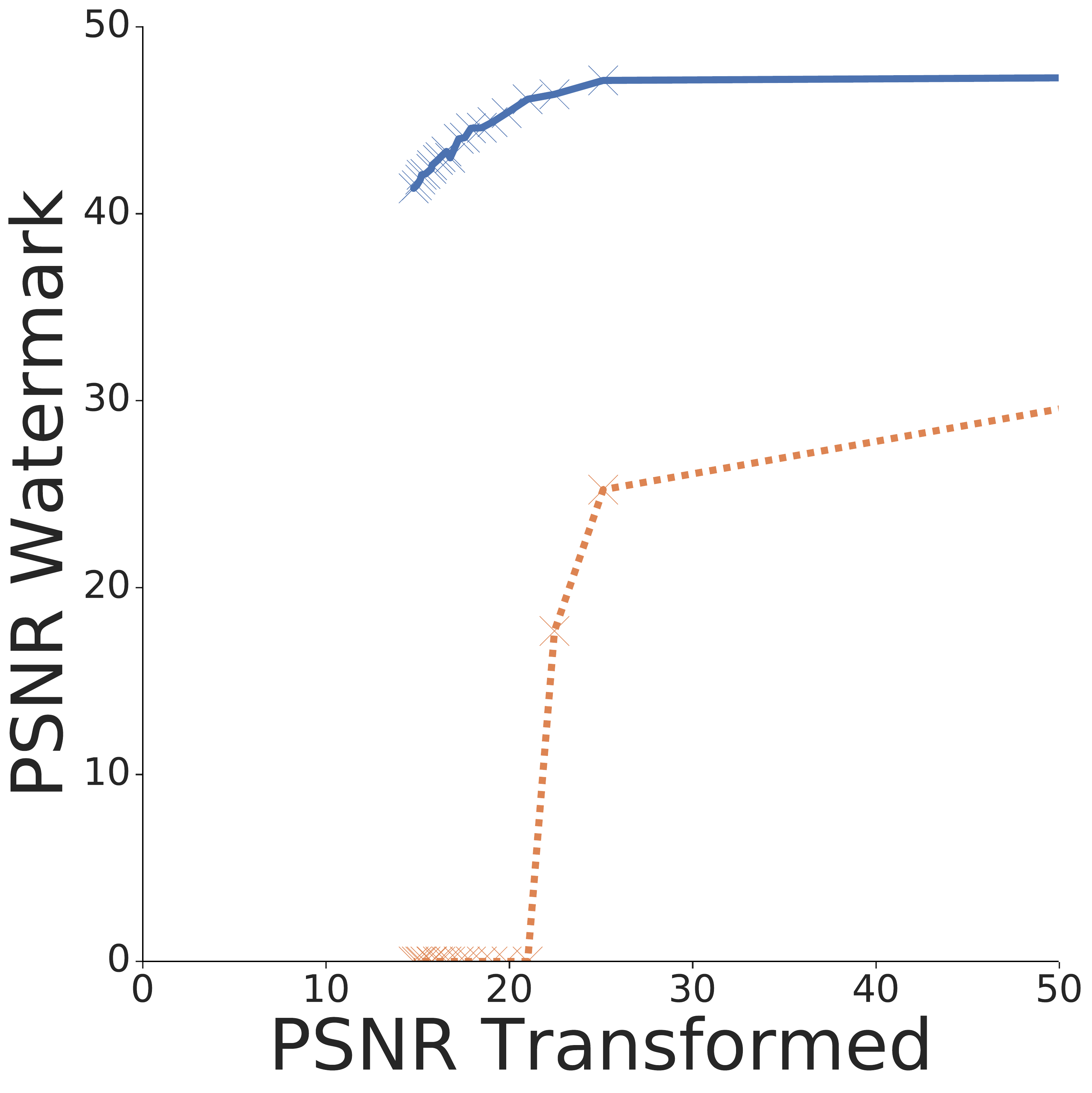}
        \label{fig:imagenet_full_acc_psnr_cropping}
        \caption{Crop.}
    \end{subfigure}
    \caption{Distortion needed for perfect provenance detection (by $f_{\text{comp}}^{\epsilon_{10}}$) against different signal quality degradation levels suffered by the attacker. We measure the amount of distortion
    the watermarking scheme must introduce in order to guarantee perfect detection accuracy
    under a transformation, and compare with the Broken Arrows zero-bit watermarking scheme.}
   \label{fig:imagenet_full_acc_complete_psnr}
\end{figure*}

\Cref{fig:imagenet_full_acc_complete_psnr} gives analogous plots to
\cref{fig:imagenet_full_acc_complete_ssim} when the measure of 
distortion introduced by both the watermark and a transformation is
Peak-Signal-to-Noise ratio (PSNR). As one may expect, results exhibited here mirror those of \cref{fig:imagenet_full_acc_complete_ssim}, \namewal
and BA are comparable under Gaussian noise transformations, while
\namewal is substantially better than BA under rotation and cropping
transformations.

\clearpage

\section{Measuring how the number of samples from the transformation distribution used during training affects resilience to attacks}
\label{sec:trans_sample_exp}

\begin{figure}[htb!]
  \centering
    \includegraphics[width=0.5\textwidth]{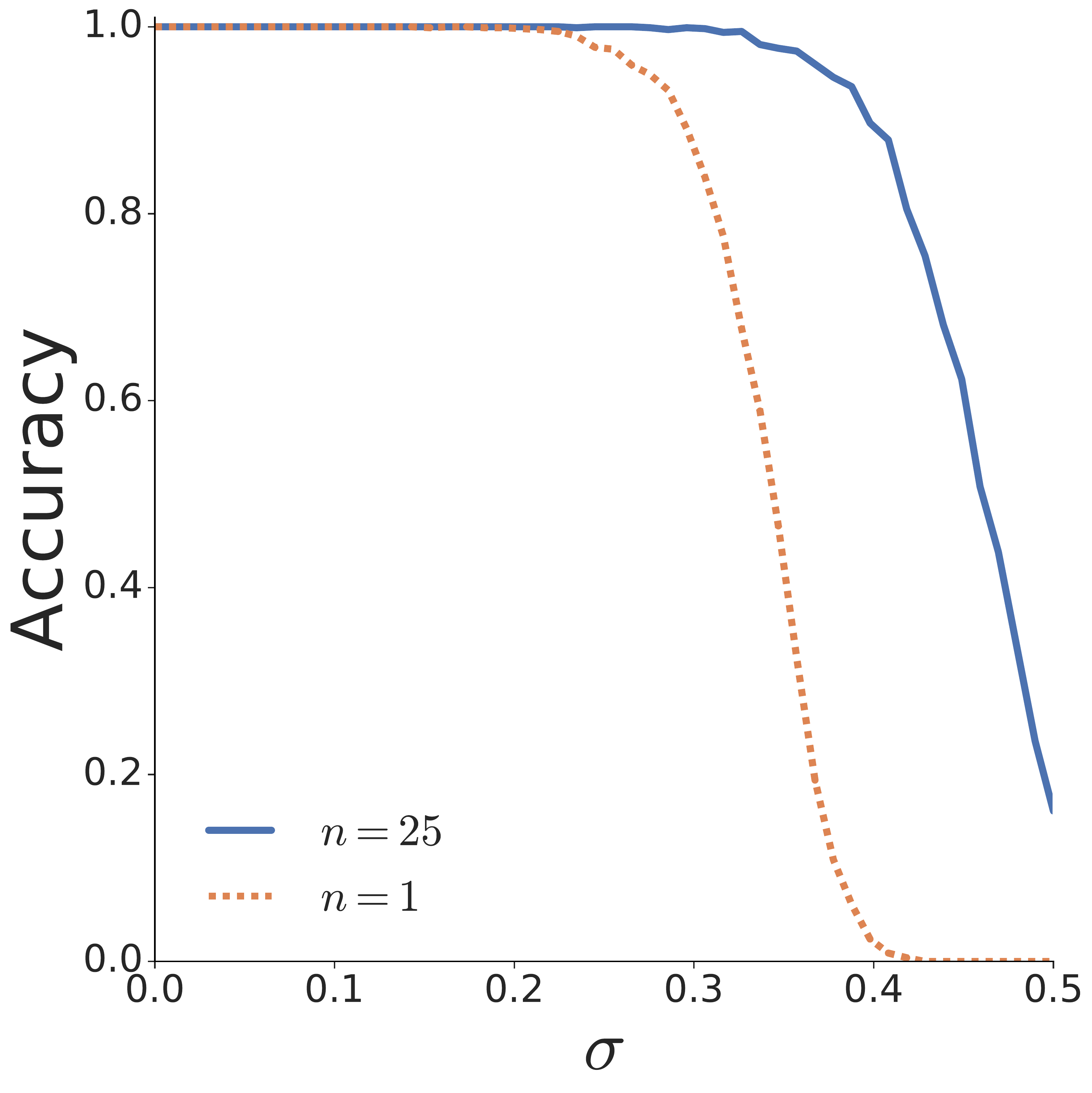}
  \caption{Comparison of \namewal trained using a single sample from the transformation distribution at each training step, and with 25 samples at each training step, for Cifar10, against a Gaussian noise attack.}
  \label{fig:eot_vs_worst_case_cifar10}
\end{figure}

Here, we evaluate the increase in resilience to 
transformations when
we optimize \namewal using \cref{eq2} with 25 samples from the transformation distribution at each step of training, as opposed to a single sample. \Cref{fig:eot_vs_worst_case_cifar10} shows
the resilience improvements of \namewal against a 
Gaussian noise transformation on the Cifar10 dataset. We evaluate test accuracy on 50,000 watermarked and non-watermarked images with various levels of Gaussian noise applied. However, the improvement in resilience
was at the expense of approximately a $5\times$ increase
in training time.

\section{Out-of-distribution transformations}
\label{sec:ood_transformations}

\begin{figure*}[htb!]
    \centering
    \begin{subfigure}[t]{0.24\textwidth}
        \centering
        \includegraphics[width=1.0\textwidth]{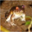}
        \label{fig:identity}
        \caption{Clean}
    \end{subfigure}
    \begin{subfigure}[t]{0.24\textwidth}
        \centering
        \includegraphics[width=1.0\textwidth]{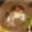}
        \label{fig:blur}
        \caption{Clean + Blur}
    \end{subfigure}
    \begin{subfigure}[t]{0.24\textwidth}
        \centering
        \includegraphics[width=1.0\textwidth]{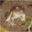}
        \label{fig:contrast}
        \caption{Clean + Contrast}
    \end{subfigure}
    \begin{subfigure}[t]{0.24\textwidth}
        \centering
        \includegraphics[width=1.0\textwidth]{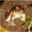}
        \label{fig:jpeg_transform}
        \caption{Clean + JPEG compression}
    \end{subfigure}
    \vskip\baselineskip
    \begin{subfigure}[t]{0.24\textwidth}
        \centering
        \includegraphics[width=1.0\textwidth]{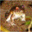}
        \label{fig:wm_identity}
        \caption{Watermarked}
    \end{subfigure}
    \begin{subfigure}[t]{0.24\textwidth}
        \centering
        \includegraphics[width=1.0\textwidth]{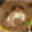}
        \label{fig:wm_blur}
        \caption{Watermarked + Blur}
    \end{subfigure}
    \begin{subfigure}[t]{0.24\textwidth}
        \centering
        \includegraphics[width=1.0\textwidth]{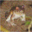}
        \label{fig:wm_contrast}
        \caption{Watermarked + Contrast}
    \end{subfigure}
    \begin{subfigure}[t]{0.24\textwidth}
        \centering
        \includegraphics[width=1.0\textwidth]{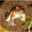}
        \label{fig:wm_jpeg_transform}
        \caption{Watermarked + JPEG compression}
    \end{subfigure}
    \caption{Examples of correctly predicted transformed images, using out-of-distribution transformations not seen by \namewal during training. The maximum watermark perturbation is set to $\epsilon=\nicefrac{20}{255}$.}
   \label{fig:ood_transformations}
\end{figure*}

So far, our evaluation has shown that \namewal is robust to post-processing transformations, if those transformations were also sampled during the training of the detector model. However, we cannot \emph{a priori} know which transformations an attacker may apply to an image in an attempt to cause a misclassification. Therefore, we are also interested in how \namewal performs against post-processing transformations that have not been used during the training process. Here, we evaluate this desired property; the robustness of \namewal when generalized to transformations that have not been seen during training.

We consider a greater number of post-processing transformations for this experiment -- in addition to Gaussian noise, rotations, cropping, and brightness changes, we include blurring, contrast changes, and JPEG compression. Below we detail the seven transformations used in out-of-distribution transformation experiments:

\begin{enumerate}
    \item \textbf{Blurring.} Blur an image with a Gaussian filter with $\sigma=3$.
    \item \textbf{Brightness.} Randomly adjust the brightness of an image by a factor of 0.2, where 0 represents no increase in brightness and 1 represents maximum whitening.
    \item \textbf{Contrast.} Increase contrast by a factor of 0.7. An enhancement factor of 0 gives a solid grey image. A factor of 1 gives the original image.
    \item \textbf{Cropping.} Randomly crop an images's height by 8 pixels and width by 8 pixels, and project back to the original image size.
    \item \textbf{Gaussian noise.} Gaussian noise is added with zero mean and standard deviation 0.1.
    \item \textbf{JPEG Compression.} Compress the image with a quality factor of 50.
    \item \textbf{Rotation.} The image is rotated by a random angle uniformly sampled from range [$\nicefrac{-\pi}{2}$,$\nicefrac{\pi}{2}$]. 
\end{enumerate}

Qualitative examples of the transformations \namewal is able to generalize to are given in \cref{fig:ood_transformations}.

For each of these seven transformations, we train a wide ResNet  detector model with a maximum watermark perturbation set to $\epsilon=\nicefrac{20}{255}$ on the Cifar10 training set, where six transformations are sampled during training and all seven are tested after training. Thus, in total, we train seven distinct watermark detectors. We then measure the accuracy of these detectors on both the six transformations observed during training, and the omitted seventh transformation not seen during training. Each accuracy value reported is the accuracy of a watermark detector on 20,000 Cifar10 test set images (10,000 watermarked and 10,000 non-watermarked images) modified with a post-processing transformation. \Cref{tab:ood_transformations} shows that \namewal is robust to transformations that have not been observed during training. We observe practically no errors on brightness, contrast, cropping, and Gaussian noise transformations. More errors are observed on transformations such as blurring, JPEG compression, and rotations, which substantially change the structure of an image.

\begin{table}[htb!]
\captionsetup{width=1.0\textwidth}
\caption{Generalization to out-of-distribution (OOD) transformations. For each of the seven transformations detailed in \cref{sec:ood_transformations}, we train a watermark detector model on the Cifar10 training set with a maximum watermark perturbation set to $\epsilon=\nicefrac{20}{255}$. We then measure the robustness of the detector to both in-distribution transformations, that have been seen during training, and the OOD transformation that was omitted from training. Robustness is measured as the accuracy of the detector on the Cifar10 test set, 10K non-watermarked test set images and 10K watermarked test set images with $\epsilon=\nicefrac{20}{255}$, when modified with a chosen transformation.}
\label{tab:ood_transformations}
\centering

\resizebox{1\columnwidth}{!}{%
\begin{tabular}{ccccccccc}

\toprule

\multicolumn{2}{c}{} & \multicolumn{7}{c}{\textbf{Accuracy on transformation (\%)}} \\ 

 \multicolumn{1}{c}{} &
 \multicolumn{1}{c}{} &
 \multicolumn{1}{c}{\rotatebox{45}{Blur}} & \multicolumn{1}{c}{\rotatebox{45}{Brightness}} &
 \multicolumn{1}{c}{\rotatebox{45}{Contrast}} & \multicolumn{1}{c}{\rotatebox{45}{Crop}} & 
 \multicolumn{1}{c}{\rotatebox{45}{Gaussian noise}} &
 \multicolumn{1}{c}{\rotatebox{45}{JPEG compression}} &
 \multicolumn{1}{c}{\rotatebox{45}{Rotation}}
  \\
  
  \cmidrule{3-9}
 
 \multirow{7}{*}{\rotatebox{90}{\textbf{OOD transformation}}} & \multicolumn{1}{l}{Blur} & 73.515 & 99.950 & 99.970 & 100 & 99.740 & 89.625 & 99.330  \\
 
  & \multicolumn{1}{l}{Brightness} & 99.965 & 99.835 & 99.990 & 99.985 & 99.875 & 99.590 & 99.835   \\
 
  & \multicolumn{1}{l}{Contrast} & 99.995 & 99.950 & 99.980 & 100 & 99.825 & 99.330 & 99.875   \\
 
  & \multicolumn{1}{l}{Crop} & 99.970 & 99.955 & 99.985 & 99.780 & 99.880 & 99.565 & 99.850   \\
 
  & \multicolumn{1}{l}{Gaussian noise} & 99.950 & 99.935 & 99.970 & 99.995 & 99.710 & 99.630 & 99.860   \\
  
   & \multicolumn{1}{l}{JPEG compression} & 100 & 99.995 & 100 & 100 & 99.960 & 67.030 & 99.950   \\
  
  & \multicolumn{1}{l}{Rotation} & 100 & 99.990 & 100 & 100 & 100 & 100 & 65.195   \\[0.5cm]

\bottomrule
\end{tabular}%
}
\end{table}

\newpage

\section{Comparison with deep learning based multi-bit watermarking methods}
\label{sec:comparison_multi_bit_related_work}

Recent work has used deep learning to develop \emph{multi-bit} watermarking schemes \citep{mun2017robust, ahmadi2018redmark, zhu2018hidden, zhang2019robust, wen2019romark, luo2020distortion}. 
As such, it is not possible to directly compare these works with ours, since they all measure accuracy in terms of the number of watermark bits recovered, while our accuracy is simply the number of successes versus failures at recognizing watermarked images. 
In addition to these different goals, \citet{zhu2018hidden, wen2019romark, luo2020distortion} also differ to our method in terms of design. 
These works introduce both an encoder model (that generates the watermark) and a detector model, that are trained in a mini-max adversarial manner, while our scheme has only one model, the detector. 
In our scheme, the encoding (watermark) is directly added to an input based on the sign of the gradient direction that causes the detector to predict the watermark label, and does not require an encoding model.

Although our aim is to develop a robust zero-bit watermarking scheme, we show here how to extend our work to multi-bit schemes, and compare with related work. In our zero-bit watermarking scheme, we parameterize the watermark detector $D$ as a neural network $f_{\theta}:\mathbb{R}^d \rightarrow \mathbb{R}$, parameterized by $\theta$. We can easily extend \namewal to a multi-bit watermarking scheme, by training $f_{\theta}:\mathbb{R}^d \rightarrow \mathbb{R}^n$, where $n$ is the number of bits to be encoded and decoded in the multi-bit scheme. Thus, instead of solving the original optimization problem:

\begin{align}
   \minimize_{\theta, \norm{\delta}_{\infty}\leq \epsilon} \ExP{\max_{T \in \mathcal{T}} \ell\br{f_\theta\br{T\br{s + \delta}}, 1} + \ell\br{f_\theta\br{T\br{s}}, 0}}{s \sim P_s}, 
\end{align}

we solve instead:

\begin{align}
   \minimize_{\theta, \norm{\delta}_{\infty}\leq \epsilon} \ExP{\max_{T \in \mathcal{T}} \ell\br{f_\theta\br{T\br{s + \delta}}, c}}{s \sim P_s, c \sim \{0, 1\}^n}, 
\end{align}

where $c$ is an $n$-bit random binary code. 

We compare with two other deep learning based multi-bit watermarking schemes, \citet{zhu2018hidden} and \citet{luo2020distortion}. 
To compare fairly, we replicate the experimental set-up in these two works. The length of the binary code, $c$, is 30, we train and evaluate the multi-bit watermark detector on the MS COCO
dataset \citep{lin2014microsoft} resized to $128\times128$, where a random selection of 3000 images are used for evaluation. We train and evaluate with two transformations used in \citet{zhu2018hidden} and \citet{luo2020distortion}: \emph{Gaussian blurring} parameterized by $\sigma$, the standard deviation for Gaussian kernel, and \emph{Dropout}. Given an image, $\hat{x}$ that has been encoded with an $n$-bit message $c$, and the original image $x$, Dropout randomly replaces $(100-p)\%$ of pixels in $\hat{x}$ with the corresponding pixels in $x$. During training we set $\sigma = 1.$ and $p=30$, as reported in \citet{luo2020distortion}. We report the average percent of correctly recovered bits by the detector. We use the hinge loss between the binary message code and output logits for multi-bit watermark training.

\begin{figure}[t]
\centering
\begin{subfigure}{.5\textwidth}
  \centering
  \includegraphics[width=\linewidth]{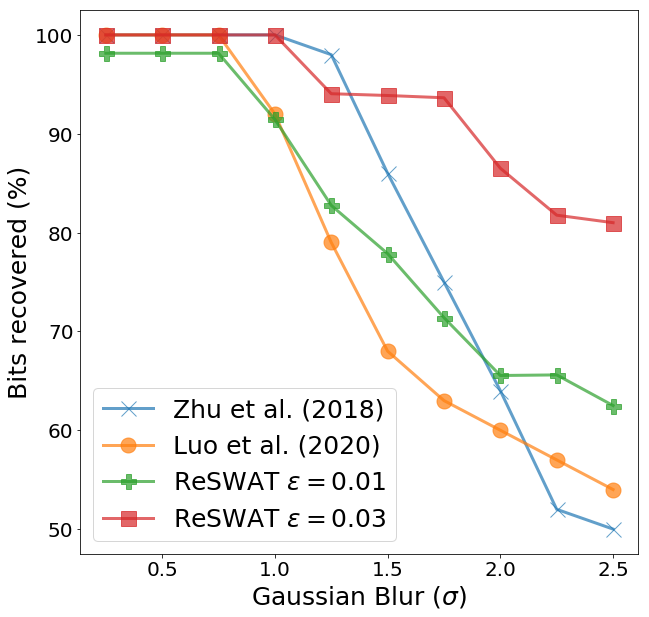}
  \caption{Gaussian blur}
  \label{fig:compare_multi_bit_blur}
\end{subfigure}%
\begin{subfigure}{.5\textwidth}
  \centering
  \includegraphics[width=\linewidth]{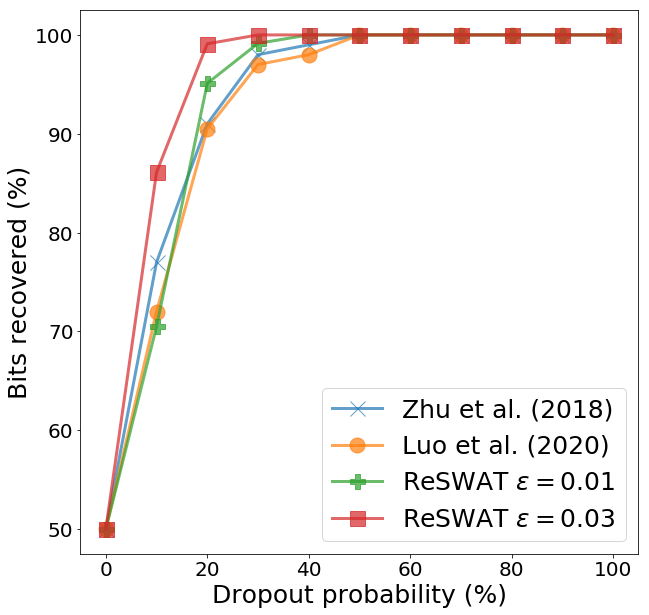}
  \caption{Dropout}
  \label{fig:compare_multi_bit_dropout}
\end{subfigure}
\caption{Robustness of multi-bit watermarking schemes against Gaussian blurring and Dropout.  The RGB-PSNR between original and encoded images are: 32.3 \citep{zhu2018hidden}, 33.7 \citep{luo2020distortion}, 40.2 (\namewal w/ $\epsilon=0.01$), 30.5 (\namewal w/ $\epsilon=0.03$).}
\label{fig:compare_multi_bit}
\end{figure}

\Cref{fig:compare_multi_bit} shows the robustness of the multi-bit \namewal with $\epsilon=0.01$ and $\epsilon=0.03$, compared with \citet{zhu2018hidden} and \citet{luo2020distortion}.
Using $\epsilon=0.03$ gives approximately equivalent PSNR rates to \citet{zhu2018hidden} and \citet{luo2020distortion} while $\epsilon=0.01$ maintains higher image fidelity. Using $\epsilon=0.01$, multi-bit \namewal is more robust than \citet{luo2020distortion} to Gaussian blurring at all $\sigma$ values, and is more robust to larger distortions ($\sigma > 2$) than \citet{zhu2018hidden}. 
Using  $\epsilon=0.03$, which gives approximately equivalent PSNR rates to \citet{zhu2018hidden} and \citet{luo2020distortion}, we substantially outperform both schemes, the average percent of correctly recovered bits does not fall below $80\%$ while \citet{zhu2018hidden} and \citet{luo2020distortion} can only correctly recover $<65\%$ of bits at $\sigma = 2.5$. Similarly, we outperform related work for any Dropout probability when $\epsilon$ is set to $0.03$.

\clearpage

\section{Robustness to targeted adversarial attacks}\label{sec: adv_attack}

\begin{figure}[t]
\centering
\begin{subfigure}{1.\textwidth}
  \centering
  \includegraphics[width=\linewidth]{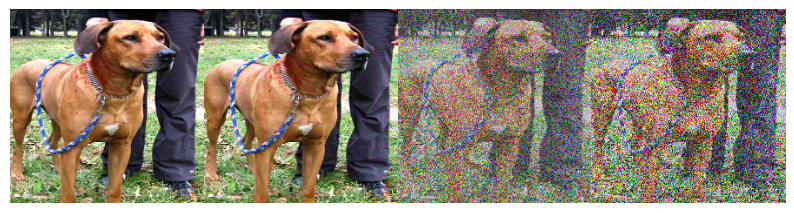}
  \caption{10 queries}
  \label{fig:sub1}
\end{subfigure}
\begin{subfigure}{1.\textwidth}
  \centering
  \includegraphics[width=\linewidth]{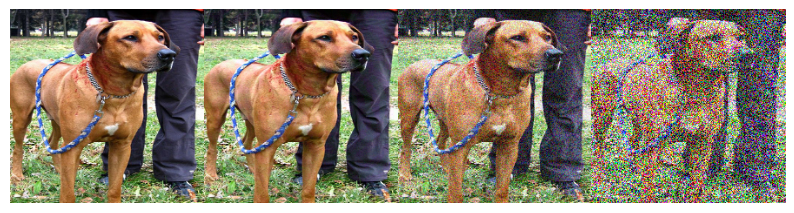}
  \caption{100 queries}
  \label{fig:sub2}
\end{subfigure}
\caption{From left to right: non-watermarked image, watermarked image, minimum amount of noise required to cause a misclassification of the non-watermarked image, minimum amount of noise required to cause a misclassification of the watermarked image. We use the the state-of-the-art black-box attack \citep{chen2020hopskipjumpattack} with either 10 or 100 queries to the \namewal detector.}
\label{fig:hsj_attack}
\end{figure}

Although throughout this work we assume that an attacker cannot observe query responses from the detector, here we show that if an attacker can observe a limited number of responses, our scheme maintains robustness to targeted attacks. 

We take the ReSWAT model trained on image transformations as described in section 4.1 and then launch black-box adversarial attacks on both watermarked and non-watermarked test set images using the state-of-the-art black-box attack \citep{chen2020hopskipjumpattack}. \Cref{fig:hsj_attack} shows that if the number of queries to the detector is limited, the black-box attack must add a large perturbation to cause a misclassification. Interestingly, as the number of allowed queries increases the required perturbation size for misclassification decreases when attacking non-watermarked images, but does not substantially decrease when attacking watermarked images.

\end{document}